\documentclass[11pt]{article}

\usepackage{ace}

\acelogos{%
  \acelogo[0.62in]{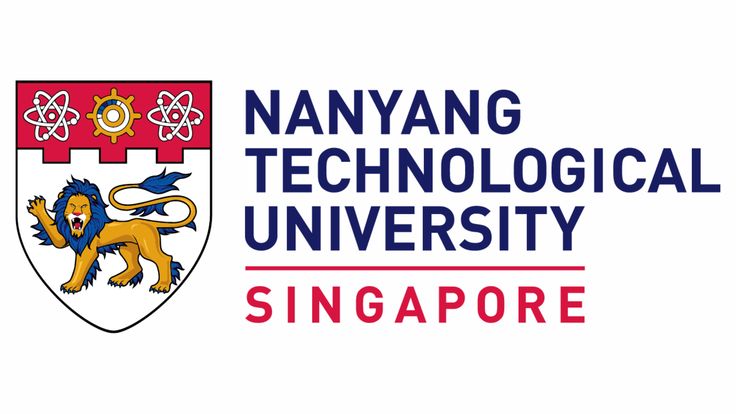}\hspace{0.9em}%
  \acelogo[0.40in]{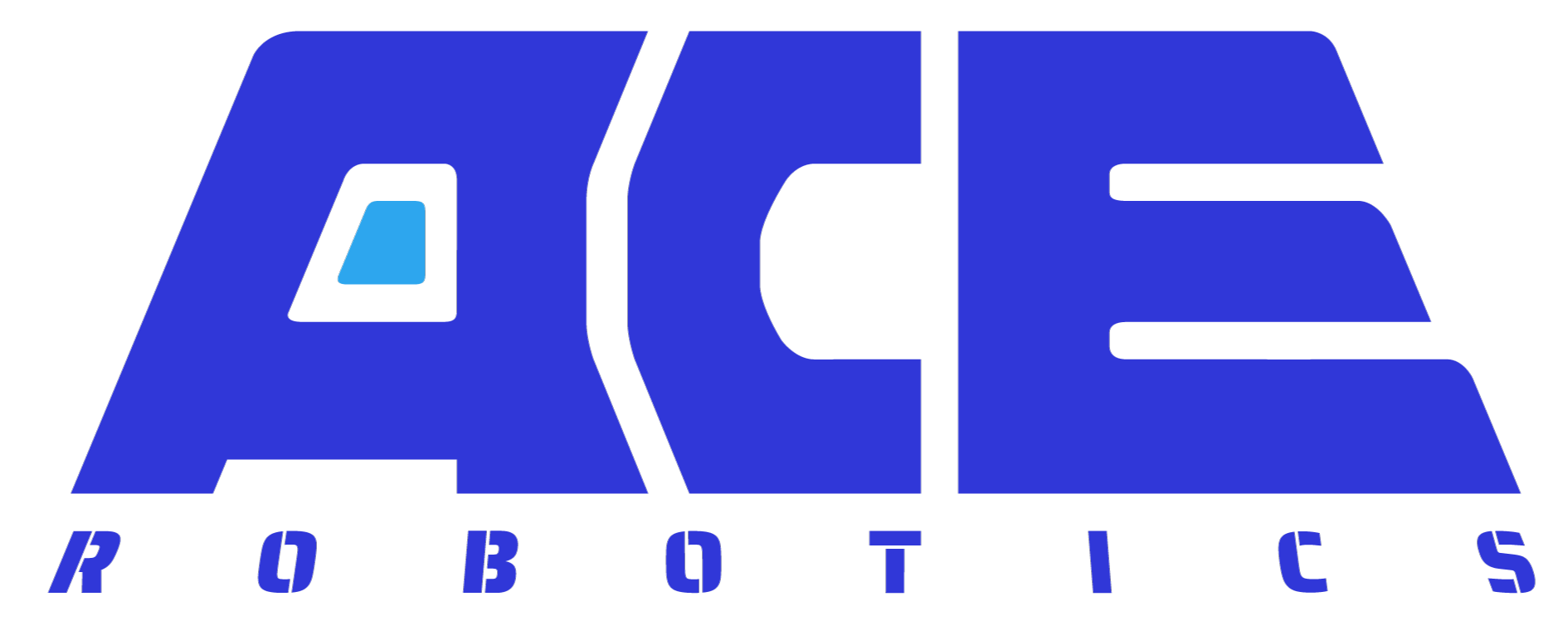}%
}

\acetitle{ACE-Data-0: Human-Centric Ambient Capture as Embodied Data Engine}

\aceauthors{%
Yukang Cao\textsuperscript{1*}, Haozhe Xie\textsuperscript{1*}, Beichen Wen\textsuperscript{2*}, Runmao Yao\textsuperscript{1}, Yinghao Liu\textsuperscript{2}, Yue Huang\textsuperscript{2}, Zhichao Liao\textsuperscript{2}, Yunxiang Wang\textsuperscript{1}, Haiheng Liu\textsuperscript{1}, Xingshun Tian\textsuperscript{2}, Dawei Su\textsuperscript{2}, Long Zhuo\textsuperscript{2}, Dacheng Tao\textsuperscript{2$\dagger$}, Xiaogang Wang\textsuperscript{2$\dagger$}, Liang Pan\textsuperscript{2$\ddagger$}, Ziwei Liu\textsuperscript{1$\ddagger$}%
}

\aceaffiliations{%
\textsuperscript{1}\,S-Lab, Nanyang Technological University \quad \textsuperscript{2}\,ACE Robotics%
}

\acenotes{%
\textsuperscript{*}\,Equal Contributions \quad \textsuperscript{$\dagger$}\,Project Advisor \quad \textsuperscript{$\ddagger$}\,Project Lead%
}

\acepagelink{https://ace-data-engine.github.io/ACE-Data-0/}

\aceteaser{%
  \noindent\includegraphics[width=\linewidth]{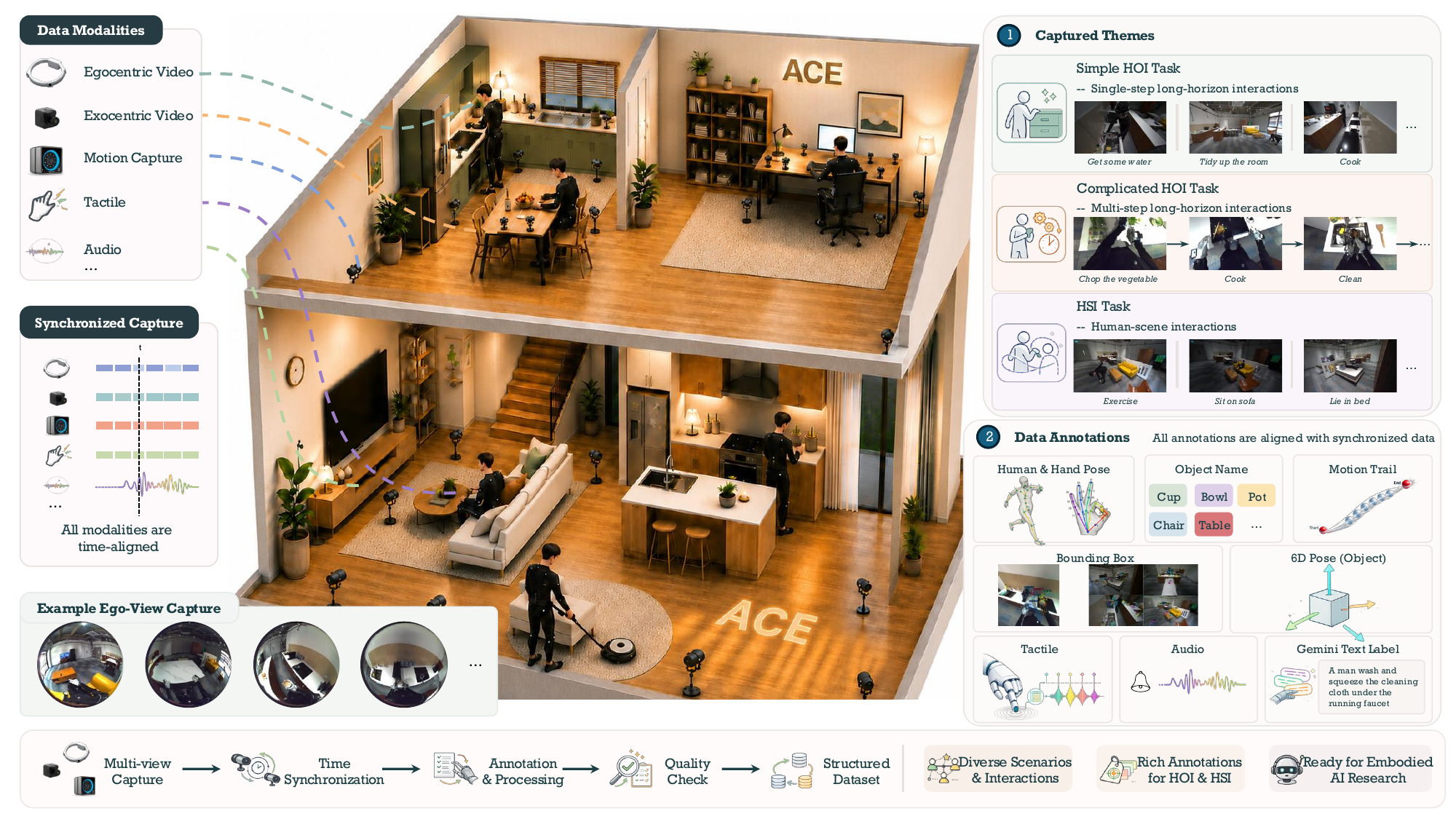}
  \captionof{figure}{%
  We introduce the Ambient Capture Engine (\textbf{ACE}), a capture system that transforms real home environments into recording studios. ACE observes household activity at two complementary scales, \textit{i.e.}, table-scale and room-scale. Via ACE, we capture and release ACE-Data-0 with synchronized multiple modalities and rich data annotations.}
  \label{fig:teaser}
}

\aceabstract{%
Embodied intelligence faces a fundamental data bottleneck. Learning to act in the physical world requires more than observing what an action looks like: \textit{models must capture how first-person perception, whole-body motion, dexterous manipulation, object state, sound, and touch evolve together as humans pursue goals over time.} Existing datasets typically fragment this experience across viewpoints, modalities, or spatial scales, leaving the full perception-action loop only partially observed.

We introduce the \textbf{A}mbient \textbf{C}apture \textbf{E}ngine (ACE), a human-centric data engine that transforms real home environments into spatially calibrated, temporally synchronized recording studios. ACE operates at two complementary scales: a \textbf{table-scale} configuration resolves fine-grained hand-object manipulation, while a \textbf{room-scale} configuration captures whole-body motion, locomotion, and interactions distributed across a furnished home. Across both settings, ACE records egocentric and multi-view exocentric video, full-body and articulated hand motion, high-fidelity object geometry and 6-DoF trajectories, multi-channel audio, and tactile signals as a unified multisensory stream.

Using ACE, we build \textbf{ACE-Data-0}, comprising 150 hours and 17M video frames across 200 task categories, performed by 50 participants in 2 environments, for a total of 75,000 interaction episodes. The dataset spans atomic manipulation, long-horizon chains of household activities, and human-scene interaction, while preserving natural behavioral variation through goal-level rather than step-by-step instructions.
We further introduce a hierarchical benchmark that progresses from \textbf{signals} to \textbf{scene components} and then to \textbf{interactions}. It encompasses tactile inference from visual observations, human motion recovery, and hand-motion reconstruction from both egocentric and exocentric perspectives.
Evaluations of state-of-the-art methods expose substantial gaps under contact, occlusion, egomotion, and long temporal horizons.

Beyond benchmarking, ACE-Data-0 provides synchronized human demonstrations with aligned perceptual, kinematic, and contact supervision, offering a scalable foundation for imitation learning, world models, vision-language-action systems, and embodied AI.%
}

\acelinks{%
  \acelinkline{Project Page}{https://ace-data-engine.github.io/ACE-Data-0/}
  \acelinkline{Hugging Face}{https://huggingface.co/datasets/ACERobotics/ACE-Data-0}
}

\begin{document}

\acemaketitle

\clearpage
\tableofcontents
\clearpage

% -----------------------
% Main text
% -----------------------
\section{Introduction}
\label{sec:intro}

Humans spend the majority of their lives in built environments, continuously interacting with the objects around them, such as opening cabinets, pouring water, folding laundry, and assembling furniture. 
Effortless as they appear, these everyday interactions embody a form of intelligence that existing models have yet to attain: the seamless coordination of perception, whole-body movement, dexterous manipulation, and physical sensing~\cite{gibson2014ecological,pfeifer2006body}.
Reproducing this intelligence is a central pursuit of embodied AI, yet unlike the AI breakthroughs that preceded it, this pursuit inherits no data. 
Language and vision models~\cite{deng2009imagenet,schuhmann2022laion,radford2021learning,brown2020languagemodelsfewshotlearners} rose on archives that humanity had spent centuries accumulating.
Physical skills, exercised without deliberation, have never been written down: how a hand closes around a cup, with what force a fragile glass is held, by what coordination of vision and balance an object is carried across a room. 
The data for embodied intelligence must therefore be built rather than found, by instrumenting everyday life itself and recording human-object interaction (HOI) as it naturally unfolds.

Constructing such a dataset, however, demands far more than pointing a camera at daily life. It requires a \textit{holistic} recording of the interaction, spanning \textit{what} the human is doing, \textit{how} the body and hands move, \textit{how} the object state evolves in response, \textit{what} audio and tactile signals the interaction produces, and \textit{how} the scene appears from both the human's first-person perspective and third-person viewports.
More importantly, such a dataset for learning embodied AI demands all of these signals to be recorded synchronously and in registration with one another, a requirement that no existing dataset satisfies.

% Specifically, existing datasets fall short in three aspects\todo{a table summary existing datasets}: 
% \textbf{(1) Fragmented modalities.} Large-scale egocentric datasets (\textit{i.e.}, Ego-Exo4D, EPIC-Kitchens, Xperience-10M \todo{More reference and cite}) offer naturalistic behavior at scale, yet provide neither ground-truth body and object motion nor synchronized third-person observation.
% Conversely, motion-captured HOI datasets, such as BEHAVE, GRAB, ARCTIC, OakInkv2, HOT3D \todo{more reference and cite}, supply accurate poses but omit the egocentric perspective.
% More importantly, audio or tactile sensing is absent from nearly all of them.
% \textbf{(2) Unnatural environments.} Physically annotated datasets \todo{reference and cite} are captured almost exclusively in laboratories, whose sparse layout eliminate precisely the occlusions, spatial constraints, and object diversity that render real homes challenging.
% \textbf{(3) Short horizons.} The vast majority of HOI clips span seconds and depict one simple movement. Genuine household activities, in contrast, are goal-directed. They may unfold over minutes or hours, chain together sub-tasks and multiple objects, and require the human to move through the scene rather than acting in place. 
% Consequently, the complete perception-action loop of everyday interaction has remained beyond the reach of existing benchmarks.

Specifically, existing datasets fall short in three aspects, as summarized in Table~\ref{tab:comparison_dataset}: 
\textbf{(1) Fragmented modalities.} Large-scale egocentric datasets (\textit{e.g.}, Ego-Exo4D~\cite{grauman2024egoexo4d}, EPIC-Kitchens~\cite{damen2020rescaling}, Xperience-10M~\cite{ropedia2026xperience10m}) offer naturalistic behavior at scale, yet provide neither ground-truth body and object motion nor synchronized third-person observation.
Conversely, motion-captured HOI datasets, such as BEHAVE~\cite{bhatnagar2022behave}, GRAB~\cite{taheri2020grab}, ARCTIC~\cite{fan2023arctic}, OakInk2~\cite{zhan2024oakink2}, and HOT3D~\cite{banerjee2025hot3d}, supply accurate poses but omit the egocentric perspective.
More importantly, audio or tactile sensing is absent from nearly all of them.
\textbf{(2) Unnatural environments.} Physically annotated datasets~\cite{brahmbhatt2020contactpose,chao2021dexycb,taheri2020grab,fan2023arctic,fu2025gigahands} are captured almost exclusively in laboratories, whose sparse layouts eliminate precisely the occlusions, spatial constraints, and object diversity that render real homes challenging.
\textbf{(3) Short horizons.} The vast majority of HOI clips span seconds and depict one simple movement~\cite{brahmbhatt2020contactpose,chao2021dexycb,kwon2021h2o,fan2023arctic}. Genuine household activities, in contrast, are goal-directed. They may unfold over minutes or hours, chain together sub-tasks and multiple objects, and require the human to move across the scene rather than acting in a single place. 
Consequently, the complete perception-action loop of everyday interaction has remained beyond the reach of existing benchmarks.

To bridge this gap, we develop the \textit{Ambient Capture Engine} (ACE), a capture system that turns real home environments into recording studios while preserving their lived-in realism. 
Fine-grained dexterous manipulation and room-scale activity impose conflicting requirements on sensor placement and coverage, so we build ACE as two complementary configurations.
The first is table-scale: densely arranged close-range cameras and high-resolution tactile sensing resolve the local details of hand-object manipulation. 
The second is room-scale: sensors span an entire furnished environment to record whole-body motion, locomotion, and interactions distributed across the scene.
While differing in sensor placement and spatial scope, both scales record synchronized egocentric video, multi-view exocentric video, optical full-body motion, and per-object 6-DoF trajectories, which are all registered into a common spatio-temporal frame. 

% Recognizing that fine-grained manipulation and room-scale activity impose conflicting requirements on sensor placement and coverage, we instantiate ACE as two complementary systems deployed at two sites: (i) \textbf{\acelocal}, a table-top configuration in which densely arranged close-range cameras and high-resolution tactile sensing resolve the local details of home-scene hand-object manipulation, and (ii) \textbf{\aceglobal}, a room-scale configuration whose sensors span an entire furnished environment to record whole-body motion, locomotion, and interactions distributed across the scene. 
% The two configurations differ in sensor placement and spatial scope rather than in sensing capability: both record synchronized egocentric video, multi-view exocentric video, optical full-body motion, per-object 6-DoF trajectories, multi-channel audio, and tactile signals, all registered into a common spatio-temporal frame.

With ACE, we collect \textbf{ACE-Data-0}: 150 hours of daily living interactions spanning 200 task categories (\textit{e.g.}, cooking, tidying, and drinking), performed by 50 participants across 2 environments, amounting to 17M video frames and 75,000 interaction episodes.
Rather than executing scripts, participants pursue goal-level instructions in their own manner: planning, hesitating, and improvising as they would at home.
Crucially, this freedom sacrifices no measurement fidelity: every moment of activity is observed simultaneously from more than 8 viewpoints and grounded in metric body, hand, and object states.
Upon these raw signals, we provide rich annotations for every sequence, including camera calibrations and synchronized timelines; full-body and hand poses and their projections onto every frame of every camera; per-object mesh models, 6-DoF poses, bounding boxes, and motion trails; and textual descriptions of the ongoing events. A large fraction of these annotations is derived automatically from the tracked physical states, instead of being estimated via existing pipelines.

Building upon ACE-Data-0, we further establish a perceptually hierarchical benchmark, comprising three tracks that advance from signals, to components, to interactions.
The first track targets \textbf{low-level signals}: 
% models localize sound sources in the scene from the recorded audio, and 
cross-modal prediction of tactile signals from visual observations.
The second track focuses on \textbf{scene components}: methods reconstruct the pose of the human body and hands, evaluated against our tracked ground-truth.
The third track evaluates the \textbf{interactions}: models estimate the hand motion during human-object interactions from egocentric and exocentric videos. Additionally, our collected paired viewpoints allow for a direct comparison between these two.
Evaluations of more than 30 representative methods expose failure modes characteristic of long-horizon home-scene data, stemming from the diversity of interactions and tasks, their complexity and interleaving, heavy occlusion, extreme viewpoints, and continual movement through the scene.
Notably, these three tracks mirror the perceptual capabilities an embodied agent must chain together: sensing contact, estimating scene state, and mastering hand-object interactions. 
The value of ACE-Data-0 to such agents, moreover, extends beyond evaluation to training itself: because our egocentric viewpoint, multi-view exocentric videos, motion trajectories, and contact-level supervision are synchronized rather than assembled from disparate sources, every training signal refers to the same physical moment, a property essential for robot learning, from imitation and policy learning to world modeling~\cite{ye2026worldactionmodelszeroshot,yang2025egovla,intelligence2025pi05visionlanguageactionmodelopenworld,su2026worldguidanceworldmodeling}.

In general, our contributions can be summarized as follows:
\begin{itemize}
    \item We present \textbf{ACE}, a ``human-centric ambient capture as embodied data engine paradigm'' realized as two complementary configurations, a table-scale setup for fine-grained local manipulation and a room-scale setup for whole-body motion in larger scenes. Within these two configurations, ACE records both temporally and spatially aligned egocentric video, multi-view exocentric video, human body and hand motion, object motion, audio, and tactile signals.
    \item We contribute \textbf{ACE-Data-0}, a large-scale, long-horizon home-scene HOI dataset comprising 17M frames and 75,000 interaction episodes, together with rich and high-quality annotations.
    \item We establish a three-level benchmark that advances from signals to components and ultimately to interactions, and provide evaluations of more than 30 state-of-the-art methods that present the open challenges for embodied perception and robot learning.
\end{itemize}

\begin{table}[t]
\centering
\caption{Comparison of ACE-Data-0 with related datasets, grouped by category.
\cmark: provided with measured (mocap-grade) ground truth; \pmark: provided but estimated, pseudo-labeled, or device-tracked; \hmark: partially provided (limited coverage or a subset); \xmark: capability absent; --: number not applicable or not reported; a \cmark in the \#Exo column denotes that exocentric video is provided without a reported camera count;
\emph{Sync}: synchronization across modality families (ego video, exo video, motion, audio, tactile); a \cmark requires at least two families, chiefly ego-exo or video-motion, recorded simultaneously and aligned, while \hmark denotes that the synchronization is conducted only among one type of viewpoint, and `--' means that only one perspective is captured. \emph{LH}: long-horizon (goal-directed activities of minutes or longer). \emph{Setup}: capture environment. In the \#Tasks column, dom., cat., scen., and skills denote domains, interaction categories, scenarios, and skills, following each dataset's own task organization.}
\label{tab:comparison_dataset}
\setlength{\tabcolsep}{4pt}%
\resizebox{\linewidth}{!}{%
\begin{tabular}{l c r r r r r c c c c c c c c l c}
\toprule
Dataset & Year & Hours & \#Frames & \#Subj & \#Obj & \#Tasks & Ego & \#Exo & Body & Hand & Obj.\ 6D & Tactile & Audio & Sync & Setup & LH \\

\catrow{Egocentric video}
Ego4D~\cite{grauman2022ego4d}
& 2022 & 3670 & -- & 931 & -- & Open
& \cmark & \xmark & \xmark & \xmark & \xmark & \xmark & \cmark & --
& In-the-wild & \cmark \\

EPIC-KITCHENS-100~\cite{damen2020rescaling}
& 2021 & 100 & 20M & 37 & -- & Open
& \cmark & \xmark & \xmark & \xmark & \xmark & \xmark & \cmark & --
& Kitchens & \cmark \\

HoloAssist~\cite{wang2023holoassist}    
& 2023 & 166 & --  & 222 & 16 & 20          %obj:16 Sync:\cmark
& \cmark & \xmark & \xmark & \pmark & \xmark & \xmark & \cmark & --
& Desktop & \cmark \\

Ego-Exo4D~\cite{grauman2024egoexo4d}
& 2024 & 1286 & -- & 740 & -- & 8 Dom.
& \cmark & 4--5 & \pmark & \xmark & \xmark & \xmark & \cmark & \hmark
& In-the-wild & \cmark \\

EgoLife~\cite{yang2025egolife}
& 2025 & 266 & -- & 6 & -- & Open
& \cmark & \cmark & \xmark & \xmark & \xmark & \xmark & \cmark & \xmark
& Shared house & \cmark \\

HD-EPIC~\cite{perrett2025hdepic}
& 2025 & 41 & 4.46M & 9 & -- & 69      %#Frames:4.46M #Tasks:69
& \cmark & \xmark & \xmark & \xmark & \pmark & \xmark & \cmark & --
& Home kitchens & \cmark \\

\catrow{Hand--object interaction}
ContactPose~\cite{brahmbhatt2020contactpose}
& 2020 & -- & 2.9M & 50 & 25 & 2
& \xmark & 3 & \xmark & \cmark & \cmark & \cmark & \xmark & \xmark
& Table-top & \xmark \\

HO-3D~\cite{hampali2020honnotate}
& 2020 & -- & 78K & 10 & 10 & --
& \xmark & 1--5 & \xmark & \cmark & \cmark & \xmark & \xmark & \xmark
& Table-top & \xmark \\

GRAB~\cite{taheri2020grab}
& 2020 & -- & 1.6M & 10 & 51 & 4
& \xmark & \xmark & \cmark & \cmark & \cmark & \pmark & \xmark & --
& Mocap lab & \xmark \\

DexYCB~\cite{chao2021dexycb}
& 2021 & -- & 582K & 10 & 20 & 1
& \xmark & 8 & \xmark & \cmark & \cmark & \xmark & \xmark & \xmark
& Table-top & \xmark \\

H$_2$O~\cite{kwon2021h2o}
& 2021 & -- & 571K & 4 & 8 & 36      %Sync:\cmark
& \cmark & 4 & \xmark & \cmark & \cmark & \xmark & \xmark & \cmark
& Table-top & \xmark \\

OakInk~\cite{yang2022oakink}
& 2022 & -- & 230K & 12 & 100 & 5
& \xmark & 4 & \xmark & \cmark & \cmark & \xmark & \xmark & \hmark
& Table-top & \xmark \\

HOI4D~\cite{liu2022hoi4d}             %Tasks:54
& 2022 & 22 & 2.4M & 9 & 800 & 54
& \cmark & \xmark & \xmark & \cmark & \cmark & \xmark & \xmark & --
& Indoor rooms & \xmark \\

ARCTIC~\cite{fan2023arctic}
& 2023 & -- & 2.1M & 10 & 11 & 2
& \cmark & 8 & \cmark & \cmark & \cmark & \xmark & \xmark & \hmark
& Mocap lab & \xmark \\

TACO~\cite{liu2024taco}
& 2024 & -- & 5.2M & 14 & 196 & 151
& \cmark & 12 & \xmark & \cmark & \cmark & \xmark & \xmark & \hmark
& Table-top & \xmark \\

HOT3D~\cite{banerjee2025hot3d}         %Audio:\xmark    %Sync:\cmark
& 2024 & 13.9 & 3.7M & 19 & 33 & Open
& \cmark & \xmark & \xmark & \cmark & \cmark & \xmark & \xmark & --
& Lab rooms & \xmark \\

OakInk2~\cite{zhan2024oakink2}         %Sync:\cmark
& 2024 & -- & 4.0M & 9 & 75 & 150
& \cmark & 3 & \pmark & \cmark & \cmark & \xmark & \xmark & \cmark
& Table-top & \pmark \\

GigaHands~\cite{fu2025gigahands}
& 2025 & 34 & 183M & 56 & 417 & Open
& \xmark & 51 & \xmark & \pmark & \pmark & \xmark & \xmark & \xmark
& Table-top & \xmark \\

\catrow{Full-body HOI, human-scene interaction, and daily motion}
BEHAVE~\cite{bhatnagar2022behave}
& 2022 & -- & 15K & 8 & 20 & --
& \xmark & 4 & \cmark & \xmark & \cmark & \xmark & \xmark & \hmark
& Lab rooms & \xmark \\

InterCap~\cite{huang2024intercap}
& 2022 & -- & 67K & 10 & 10 & --
& \xmark & 6 & \cmark & \pmark & \cmark & \xmark & \xmark & \hmark
& Lab room & \xmark \\

CHAIRS~\cite{jiang2023chairs}
& 2022 & 17.3 & -- & 46 & 81 & 32
& \xmark & 4 & \cmark & \cmark & \cmark & \xmark & \xmark & \hmark
& Lab room & -- \\

EgoBody~\cite{zhang2022egobody}
& 2022 & -- & 220K & 36 & -- & 5 Cat.
& \cmark & 3--5 & \pmark & \pmark & \xmark & \xmark & \xmark & \cmark
& Indoor rooms & \xmark \\

Aria Digital Twin~\cite{pan2023ariadigitaltwin}
& 2023 & 6.6 & -- & -- & 398 & Open
& \cmark & \xmark & \pmark & \xmark & \cmark & \xmark & \cmark & --
& Apartment, office & \xmark \\

OMOMO~\cite{li2023omomo}
& 2023 & 10 & -- & 17 & 15 & --
& \xmark & \xmark & \cmark & \xmark & \cmark & \xmark & \xmark & --
& Lab room & \xmark \\

HIMO~\cite{lv2024himo}
& 2024 & -- & 4.1M & 34 & 53 & --
& \cmark & \xmark & \cmark & \cmark & \cmark & \xmark & \xmark & --
& Mocap lab & \xmark \\

TRUMANS~\cite{jiang2024trumans}
& 2024 & 15 & 1.6M & 7 & 20 & --
& \xmark & \xmark & \cmark & \xmark & \pmark & \xmark & \xmark & --
& Scene mockups & \xmark \\

ParaHome~\cite{kim2024parahome}
& 2024 & 8.1 & -- & 38 & 22 & --
& \xmark & 70 & \cmark & \cmark & \cmark & \xmark & \xmark & \cmark
& Home room & \cmark \\

Nymeria~\cite{ma2024nymeria}
& 2024 & 300 & -- & 264 & -- & 20 Scen.
& \cmark & \cmark & \cmark & \xmark & \xmark & \xmark & \cmark & \cmark
& In-the-wild & \cmark \\

HuMoTo~\cite{lu2025humoto}
& 2025 & 2.2 & 236K & 1 & 63 & --
& \xmark & \xmark & \cmark & \cmark & \cmark & \xmark & \xmark & --
& Scene & \xmark \\

\catrow{Robot data and human demonstrations for robots}
BridgeData V2~\cite{walke2023bridgedatav2}
& 2023 & -- & 60K Traj & -- & 100+ & 13 Skills
& \xmark & \cmark & \xmark & \xmark & \xmark & \xmark & \xmark & --
& Table-top & \xmark \\

RH20T~\cite{fang2023rh20t}
& 2023 & -- & 110K Traj & -- & -- & 147
& \cmark & \cmark & \xmark & \xmark & \xmark & \pmark & \cmark & \cmark
& Table-top & \xmark \\

DexCap~\cite{wang2024dexcap}
& 2024 & -- & -- & -- & -- & --
& \xmark & \cmark & \xmark & \cmark & \xmark & \xmark & \xmark & \hmark
& Table-top & \xmark \\

DROID~\cite{khazatsky2024droid}
& 2024 & 350 & 76K Traj & -- & -- & 86
& \xmark & 3 & \xmark & \xmark & \xmark & \xmark & \xmark & \hmark
& In-the-wild & \xmark \\

AgiBot World~\cite{agibot2025colosseo}
& 2025 & 2976.4 & 1M Traj & -- & 3000+ & 217
& \xmark & \cmark & \xmark & \xmark & \xmark & \cmark & \xmark & \xmark
& Staged scenes & \cmark \\

EgoDex~\cite{hoque2025egodex}
& 2025 & 829 & 90M & -- & -- & 194
& \cmark & \xmark & \xmark & \pmark & \xmark & \xmark & \xmark & --
& Table-top & \xmark \\

Galaxea Open-World~\cite{jiang2025galaxea}
& 2025 & 500 & 100K Traj & -- & 1600 & 150
& \xmark & \cmark & \xmark & \xmark & \xmark & \xmark & \xmark & \hmark
& Open-world & \pmark \\

EgoScale~\cite{zheng2026egoscale}
& 2026 & 20,854 & -- & -- & 43,237 & 6015
& \cmark & \xmark & -- & -- & -- & \xmark & \xmark & --
& Table-top & \xmark \\

Xperience-10M~\cite{ropedia2026xperience10m}
& 2026 & 10,000 & 2.88B & -- & -- & --
& \cmark & \xmark & \pmark & \pmark & \xmark & \xmark & \cmark & --
& In-the-wild & \xmark \\

\catrow{Ours}
\textbf{ACE-Data-0}
& 2026 & \textbf{150} & \textbf{17M} & \textbf{50}
& \textbf{50} & \textbf{200}
& \textbf{\cmark} & \textbf{8}
& \textbf{\cmark} & \textbf{\cmark} & \textbf{\cmark}
& \textbf{\cmark} & \textbf{\cmark} & \textbf{\cmark}
& \textbf{Home (Room + Table-top)} & \textbf{\cmark} \\

\bottomrule
\end{tabular}%
}
\end{table}

\section{Related Work}

% \subsection{Multi-modal Datasets \& Benchmarks}

% \subsection{Egocentric Datasets \& Benchmarks}

% \subsection{Long-horizon Datasets \& Benchmarks}

\subsection{Multi-modal Datasets \& Benchmarks}

Everyday interaction is inherently multi-modal. A single physical event is reflected simultaneously in visual appearance, body and hand motion, changes in object state, acoustic cues, and physical contact. These signals are complementary rather than interchangeable: vision captures what is externally observable, kinematics describes how the human and objects evolve, audio reveals impacts and state transitions, and touch provides direct evidence of contact and force. Accordingly, datasets for embodied perception have progressively moved beyond isolated RGB observations toward richer combinations of geometry, motion, contact, and semantic annotations.

Early physically grounded datasets primarily focused on the local geometry of grasping and manipulation. ContactPose~\cite{brahmbhatt2020contactpose} associates 3D hand pose with object-surface contact over approximately 2.9M multi-view RGB-D images from 50 participants and 25 objects, while DexYCB~\cite{chao2021dexycb} provides 582K frames from eight synchronized RGB-D cameras with MANO hand~\cite{Romero_2017} annotations and YCB object poses. H$_2$O~\cite{kwon2021h2o} extends this setting to bimanual interaction through synchronized first- and third-person RGB-D observations, 3D hand keypoints, and object poses. H$_2$O-3D~\cite{hampali2022keypoint} further contributes more than 76K annotated images for challenging two-hand-object pose estimation under severe self-occlusion. At a larger semantic scale, HOI4D~\cite{liu2022hoi4d} contains 2.4M egocentric RGB-D frames involving roughly 800 object instances and supports category-level object pose tracking, action segmentation, and dynamic 4D interaction understanding. A complementary line of work expands the captured state from the hands to the full human-object system. GRAB~\cite{taheri2020grab} records detailed whole-body, hand, face, and object motion for interactions with 51 objects. BEHAVE~\cite{bhatnagar2022behave} and InterCap~\cite{huang2024intercap} jointly recover humans and rigid objects from multi-view RGB-D observations, enabling the study of body-scale coordination, contact, and joint reconstruction.

More recent datasets increasingly emphasize dexterity, compositionality, and task structure. ARCTIC~\cite{fan2023arctic} captures approximately 2.1M images of bimanual interaction with articulated objects from one egocentric and eight allocentric views, grounded by optical motion capture. TACO~\cite{liu2024taco} organizes 5.2M frames around compositional tool-action-object relationships, whereas OakInk2~\cite{zhan2024oakink2} represents complex bimanual activity through a hierarchy of affordances, motion primitives, and task-level structures. At a larger scale, GigaHands~\cite{fu2025gigahands} collects roughly 183M frames from 56 participants and 417 objects using a 51-camera system, substantially increasing the diversity of bimanual activities and language annotations. HOT3D~\cite{banerjee2025hot3d} provides approximately 3.7M egocentric multi-view images from 19 participants and 33 scanned objects, together with metric trajectories for hands, objects, cameras, and gaze. Collectively, these datasets have substantially advanced hand pose estimation~\cite{pavlakos2023hamer,zhang2025hawor,ye2025haptic}, human-object reconstruction~\cite{huang2024intercap,chen2025hort}, contact reasoning~\cite{brahmbhatt2020contactpose,fan2023arctic,muralidhar2025physic},
and interaction understanding~\cite{liu2024taco,fu2025gigahands,banerjee2025hot3d}. Their sensing configurations, however, are typically optimized for a particular spatial scale or source of supervision. Hand-centric datasets resolve detailed local geometry but usually confine activity to a compact workspace, whereas whole-body datasets preserve global coordination but may offer less detailed hand articulation or omit the actor's first-person view. Existing multi-modal resources also predominantly combine RGB, depth, pose, geometry, gaze, and language~\cite{cao2025reconstructing4dspatialintelligence}. Audio is rarely aligned with metric human and object motion at the interaction level, and tactile sensing is scarcer still, despite directly revealing contact onset, pressure, and slip. Moreover, annotations may combine direct measurement with offline fitting or model-based reconstruction, limiting their uniformity under sustained occlusion, rapid motion, or long-duration activity. High-fidelity capture is therefore often achieved in controlled spaces that simplify the clutter, furniture occlusion, and spatial constraints of domestic interaction.

ACE-Data-0 complements these efforts by treating synchronized multisensory grounding as the central unit of data. The table-scale configuration resolves fine-grained hand-object manipulation, whereas the room-scale configuration captures full-body motion and interactions distributed across a furnished domestic environment. Both configurations provide egocentric video, multi-view exocentric video, full-body and articulated hand motion, object 6-DoF trajectories, audio, and tactile signals through a shared synchronization and calibration pipeline. The modalities therefore describe the same physical event on a common timeline and in a common spatial frame, rather than serving as independently produced annotations. This unified grounding supports a coherent benchmark progression from low-level signal inference, through scene component recovery, to interaction understanding.

\subsection{Egocentric Datasets \& Benchmarks}

Egocentric video observes activity from the viewpoint of the acting subject rather than that of an external camera. This perspective naturally emphasizes action-relevant objects, hand proximity, gaze allocation, and the immediate visual consequences of self-motion, making it particularly relevant to embodied agents. At the same time, it introduces distinctive challenges: the camera moves continuously, motion blur is frequent, the hands often occlude manipulated objects, and much of the actor's body remains outside the field of view.

Large-scale datasets first established the breadth and naturalism of egocentric perception. Ego4D~\cite{grauman2022ego4d} collects 3,670 hours of unscripted first-person video from 931 camera wearers across 74 locations, with benchmarks spanning episodic memory, forecasting, social interaction, manipulation, and audio-visual understanding. HoloAssist~\cite{wang2023holoassist} focuses on interactive task assistance, recording 166 hours of instructor-performer activity with synchronized RGB-D, gaze, head and hand pose, IMU, audio, and dialogue; these signals support mistake detection, intervention prediction, and hand-motion forecasting. Ego-Exo4D~\cite{grauman2024egoexo4d} further connects first- and third-person perspectives through 1,286 hours of synchronized skilled activity from 740 participants across 123 scenarios, together with audio, gaze, language, camera poses, and 3D scene information. Its paired-view design enables cross-view activity understanding, proficiency estimation, and pose recovery on the same underlying events.

Manipulation-oriented datasets trade some behavioral breadth for stronger physical supervision. H$_2$O~\cite{kwon2021h2o} and HOI4D~\cite{liu2022hoi4d} associate first-person RGB-D observations with hand and object states, while HOT3D~\cite{banerjee2025hot3d} provides calibrated egocentric multi-view recordings with metric trajectories for hands, objects, cameras, and gaze. EgoDex~\cite{hoque2025egodex} scales human dexterous demonstrations to 829 hours across 194 task types using wearable hand tracking, creating a large source of action-relevant human manipulation data. PH2D~\cite{qiu2025humanoid} further connects human first-person demonstrations with humanoid policy learning through a robot-compatible action representation. A complementary source of first-person experience comes from robot demonstration data. Fourier ActionNet~\cite{fourier2025actionnet} records more than 30K teleoperated trajectories, totaling approximately 140 hours of dexterous bimanual manipulation across multiple humanoid platforms; it pairs observations from the robot's first-person cameras with language annotations and executable robot actions. Human egocentric datasets and robot demonstrations therefore provide complementary forms of supervision: the former preserve natural human strategies and embodiment-independent behavior, whereas the latter provide direct action labels in the target robotic embodiment.

Recent human-to-robot learning methods increasingly combine these complementary data sources. DexMV~\cite{qin2022dexmv}, EgoMimic~\cite{kareer2025egomimic}, and EgoBridge~\cite{punamiya2025egobridge} explicitly align or adapt human demonstrations to robotic embodiments. EgoVLA~\cite{yang2025egovla}, UniVLA~\cite{bu2025univla}, and In-N-On~\cite{cai2025innon} instead learn transferable action representations from mixtures of human videos and robot demonstrations. EgoScale~\cite{zheng2026egoscale} and Emergence of Human to Robot Transfer~\cite{kareer2025emergence} further suggest that transfer benefits from increasing the scale and diversity of human experience, robot tasks, and embodiments. These developments reinforce the value of first-person data, but also expose a recurring trade-off between behavioral diversity and physical completeness. Large naturalistic collections capture broad activity distributions and extended temporal context, yet do not continuously measure full-body state, object trajectories, and physical contact throughout every sequence. More instrumented datasets provide stronger hand or object supervision, but often remain local in spatial scope and omit some combination of synchronized room-scale external views, audio, or tactile sensing.

ACE-Data-0 addresses this gap by embedding egocentric observation within a shared physical representation of the surrounding event. The wearable cameras preserve the close-range perspective available to an embodied agent, while synchronized exocentric views recover body and scene context that is frequently invisible from the first-person view. Both perspectives are registered with measured headset, human, and object motion, enabling direct comparison and fusion under the same physical ground truth. Audio and tactile streams further connect visual observation to the acoustic and contact consequences of the same interaction.

\subsection{Long-horizon Datasets \& Benchmarks}

Long-horizon interaction~\cite{zhan2024oakink2,kim2024parahome,ahn2022saycan,torne2026mem} is defined not merely by recording duration, but by persistent dependencies among actions, objects, and scene states. In household tasks, earlier decisions alter later possibilities, objects move in and out of relevance, and sub-tasks may be interrupted, reordered, or resumed elsewhere. Locomotion and manipulation are likewise interdependent: the actor must move through the environment while retaining object locations, intermediate states, and the remaining goal. These properties require models to maintain task and scene memory rather than treating an activity as a sequence of independent atomic actions.

Several human-centered datasets have begun to preserve this richer temporal structure. OakInk2~\cite{zhan2024oakink2} organizes bimanual manipulation into hierarchical task components rather than isolated grasps, supporting the study of affordances, motion primitives, and complex task completion. OMOMO~\cite{li2023omomo} models the temporal coupling between object motion and full-body human motion over extended interactions, demonstrating how object trajectories constrain human behavior beyond a single contact event. ParaHome~\cite{kim2024parahome} records 207 household activity sequences, totaling approximately 486 minutes, from 38 participants using 70 synchronized RGB cameras and wearable motion capture; it emphasizes continuous, concurrent, and multi-object interaction in domestic settings. HuMoTo~\cite{lu2025humoto} contains 735 motion-capture sequences involving 63 objects and 72 articulated parts, with task designs that emphasize purposeful progression and coherent multi-object activity. A parallel line of work synthesizes extended human motion under semantic, kinematic, collective, or musical conditioning~\cite{gu2026motok,li2026infinitedance,cao2024avatargozeroshot4dhumanobject}, which likewise depends on motion data that remains coherent over long durations. Together, these datasets mark an important transition from atomic HOI clips toward scene evolution and task-level temporal structure.

Robot-learning datasets address the same challenge from the control side. Mobile ALOHA~\cite{fu2024mobilealoha} combines navigation and bimanual manipulation in 290 whole-body teleoperation demonstrations across seven tasks. AgiBot World~\cite{agibot2025colosseo} provides over one million trajectories across 217 tasks, while Galaxea Open-World~\cite{jiang2025galaxea} contains approximately 100K trajectories spanning 150 tasks. RoboCOIN~\cite{wu2025robocoin} includes roughly 180K demonstrations across 421 tasks and 15 robotic embodiments. RoboMIND 2.0~\cite{hou2025robomind2} further scales to approximately 310K trajectories and 739 tasks across six embodiments, with RGB-D observations, mobile manipulation, digital-twin assets, and a subset of tactile episodes. Real-to-sim reconstruction offers a complementary route to scaling such data: HSImul3R~\cite{cao2026hsimul3rphysicsintheloopreconstructionsimulationready} reconstructs physical scenes with physics in the loop, producing simulation-ready environments in which robot trajectories can be generated at low cost. Recent methods make the computational demands of long-horizon behavior explicit. SayCan~\cite{ahn2022saycan} connects high-level task reasoning with executable skills, MEM~\cite{torne2026mem} introduces multi-scale embodied memory, and WholeBodyVLA~\cite{jiang2026wholebodyvla} integrates vision-language-action learning with whole-body loco-manipulation. DynamicVLA~\cite{xie2026dynamicvla} instead targets manipulation of moving objects, where temporal anticipation and closed-loop adaptation become necessary. EMMA~\cite{zhu2025emma} and HoMMI~\cite{xu2026hommi} further use egocentric human demonstrations to reduce the cost of collecting mobile-robot trajectories.

Human-centered and robot-centered datasets nevertheless provide different forms of supervision. Human recordings preserve natural task decomposition, flexible sub-task ordering, hesitation, and recovery, while remaining largely independent of any particular robot embodiment. Robot datasets provide executable controls and precisely aligned observations, but their trajectories are tied to specific kinematics, sensors, and teleoperation interfaces. Few existing resources combine the natural behavioral structure of human demonstrations with continuous measurement of corresponding body, object, scene, and contact states. This separation also limits diagnosis: long-horizon robot benchmarks often emphasize final task success, whereas human activity datasets commonly focus on recognition, segmentation, or motion reconstruction. Final success alone cannot reveal whether failure originated from missed contact, inaccurate object-state estimation, lost task context, incorrect sub-task ordering, or poor motion execution.

ACE-Data-0 is designed to expose these intermediate structures in goal-directed household activity. Participants receive goal-level instructions rather than fixed atomic scripts, allowing object choice, task ordering, movement paths, hesitation, and recovery to emerge naturally. Because visual observations, human and object motion, audio, and tactile signals remain synchronized throughout each task, the dataset bridges geometric HOI benchmarks, long-form egocentric video, and robot trajectory corpora, supporting analysis of not only what occurs, but also the signals, states, and interaction dynamics through which a long-horizon task is carried out.
\section{Ambient Capture Engine}

In this section, we present the Ambient Capture Engine (ACE) designed for collecting synchronized multi-modal data.
Specifically, we first provide an overview of the system and its two scales (Section~\ref{sec:ace_overview}), and then describe the capture systems in detail, covering the home environments in which they are deployed and the sensors placed within them (Section~\ref{sec:ace_camera}).

% we first provide an overview of the Ambient Capture Engine and its two instantiations in Section~\ref{sec:ace_overview}, and the describe the capture systems, including the environments in which they are deployed and their sensor configurations (Section~\ref{sec:ace_camera}).
% Section~\ref{sec:ace_pipe} presents the data acquisition pipeline, covering the capture workflow, multi-modal recording, sensor synchronization, and calibration.
% In Section~\ref{sec:ace_data}, we illustrate the data collection process, including task design, capture settings, dataset statistics, and modalities.
% Finally, Section~\ref{sec:ace_annotation} introduces the annotations provided with the dataset and the pipeline used to produce them.

\subsection{System Overview}
\label{sec:ace_overview}

\begin{figure}[t]
    \centering
    \begin{subfigure}[t]{0.46\linewidth}
        \includegraphics[width=\linewidth]{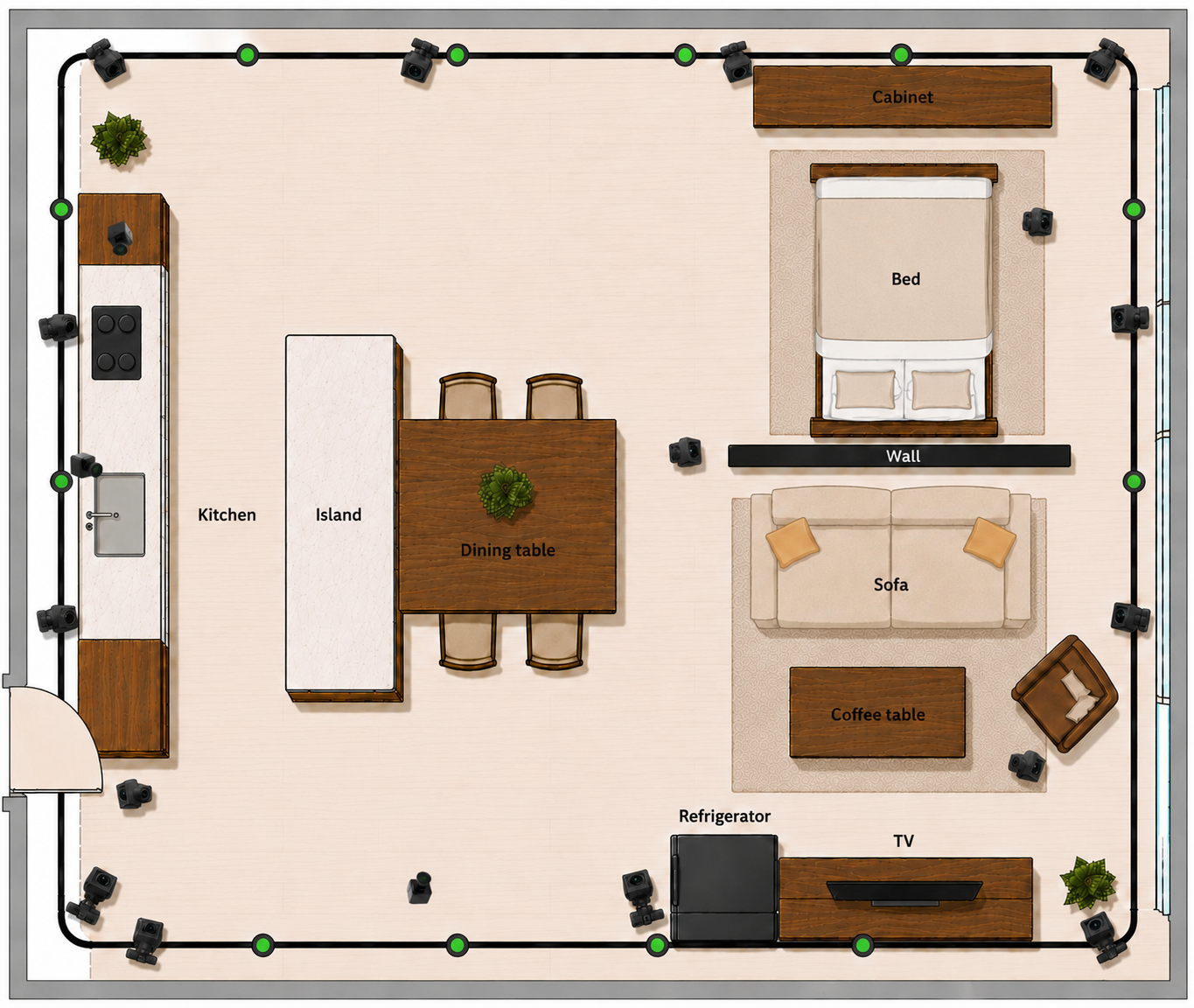}
        \caption{Home floor plan and camera setup of site I.}
        \label{fig:ace-r-setup}
    \end{subfigure}
    \hfill
    \begin{subfigure}[t]{0.53\linewidth}
        \includegraphics[width=\linewidth]{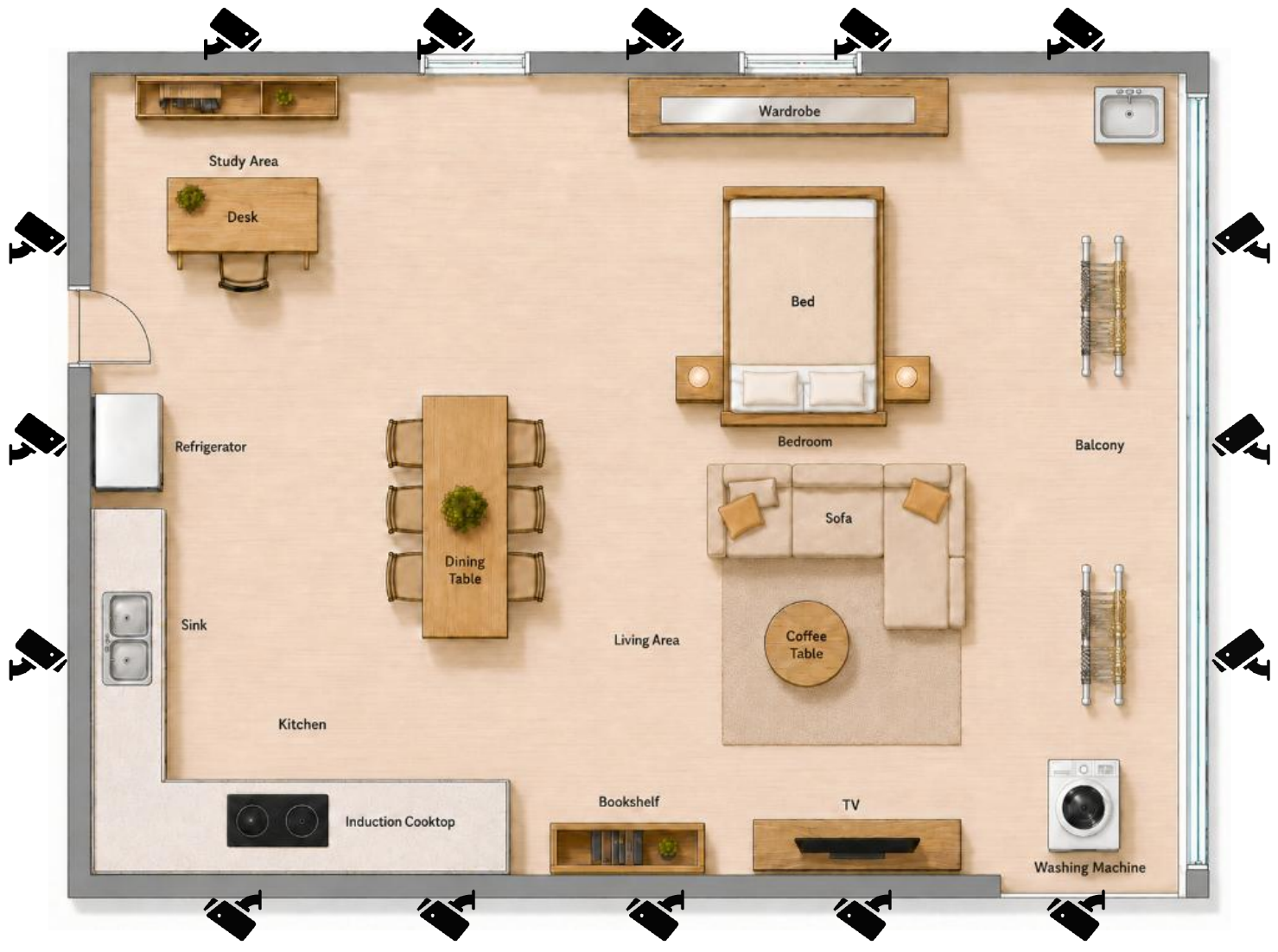}
        \caption{Home floor plan and camera setup of site II.}
        \label{fig:ace-t-setup}
    \end{subfigure}
    \caption{\textbf{Overview of the Ambient Capture Engine.} We design two complementary home scenes and capture at two spatial scales, with the equipment installed accordingly.}
    \label{fig:ace-setup}
\end{figure}
In Fig.~\ref{fig:ace-setup}, we provide the overall architecture of ACE. 
Recording everyday interaction holistically places three obligations on the capture system: it must capture every important signal an interaction may produce, keep those signals synchronized and registered in a common frame, and do all of these in environments that remain believably domestic.
No single installation, however, can satisfy the first obligation at every spatial scale: resolving finger-object contact calls for close-range, densely placed sensors, whereas following locomotion across a scene calls for wide baselines and full-room coverage.
Rather than compromise between the two, we build ACE at two spatial scales, table-scale and room-scale, and currently deploy it across two sites.
The table-scale configuration concentrates on close-range cameras, optical motion capture, and tactile gloves, targeting fine-grained dexterous hand-object manipulation.
The room-scale configuration spreads the same sensing suite over a fully furnished apartment, with wide-baseline cameras and optical motion capture covering the full activity area, targeting global motion and interactions distributed across the scene.
Both systems share a common acquisition and annotation pipeline: all sensor streams are temporally synchronized, in hardware where sensors permit and in software otherwise (Section~\ref{subsec:synchronize}), registered into a shared world frame (Section~\ref{subsec:calibration}), and enriched with annotations (Section~\ref{sec:ace_annotation}).
The result is a unified corpus in which every frame of every camera can be related to the ground-truth state of the human body, every tracked object, and the accompanying audio and tactile signals; this property underpins the benchmark in Section~\ref{sec:benchmark}.

\subsection{Capture System}
\label{sec:ace_camera}
Having sketched the architecture, we now zoom in on its physical form: first the environments that host the interactions, and then the sensors that observe them.

\subsubsection{Environment Setup}
A central design principle of ACE is to preserve the visual and physical realism of a lived-in home while permitting dense sensor coverage. The clutter, furniture, and spatial constraints that laboratory capture removes are precisely what make home-scene interaction difficult, so our environments deliberately keep them.
The table-scale configuration is built around a work desk. It covers 30 square meters and is populated with over 25 interactable object instances from more than 8 categories, such as knives, bowls, and food containers.
The workspace is instrumented with $8$ exocentric RGB cameras rigidly mounted on stands at close range (0.3--0.5 m), providing full coverage of the manipulation area and yielding multiple synchronized views at sub-millimeter effective resolution.
An optical motion capture system with 16 OptiTrack cameras, mounted on a shared truss, spans a tracking volume that covers the entire workspace.
An overview of the table-scale configuration is presented in Fig.~\ref{fig:ace-t-setup}.
On the other hand, the room-scale configuration comprises a fully furnished apartment of approximately $200$ square meters, including a kitchen with a kitchen island, a dining area, a living room, and a bedroom, populated with over 25 object instances.
A truss suspended from the ceiling and spanning the apartment carries both the RGB cameras and the optical motion-capture system: $8$ exocentric RGB cameras hang from the truss, each on an adjustable extension pole that fine-tunes its height and viewing angle, so that any point in the activity area remains visible from at least 4 views even under furniture occlusion; 
$12$ OptiTrack cameras are mounted on the same truss, with their positions and orientations tuned so that the full apartment lies within a single tracking volume. An overview of the room-scale configuration is presented in Fig.~\ref{fig:ace-r-setup}.

\begin{table}[t]
\centering
\caption{Capture hardware of ACE for synchronized multi-modal recording. Quantities are the totals for both configurations combined; a dash marks an entry that does not apply to that device.}
\label{tab:hardware_inventory}
\footnotesize
\setlength{\tabcolsep}{5pt}
\renewcommand{\arraystretch}{1.2}
% Resolution and rate share one column so that Notes gets the width it needs,
% and Notes is set ragged right: justifying a column this narrow forces the
% hyphenation to break words badly.
\begin{tabularx}{\linewidth}{@{}l l c l >{\raggedright\arraybackslash}X@{}}
\toprule
Device & Role & Qty & Resolution / rate & Notes \\
\midrule
OptiTrack PrimeX 22 & Optical motion capture & 28 & $2048\times1088$, 60\,Hz & IR tracking: 41 body markers, objects, ego rig \\
ZED One & Exocentric RGB capture & 8 & $1920\times1080$, 30\,FPS & GMSL2 to a single Jetson Orin host; shared frame trigger \\
GoPro & Exocentric RGB capture & 8 & $1920\times1080$, 30\,FPS & Rigidly mounted on stands; audio-triggered recording \\
ACE-Ego-Head-V02 Lite & Egocentric capture & 4 cameras & $4\times1088\times1280$, 20\,FPS & One headset: front/back fisheye pairs, IMU, 5 markers \\
Manus & Hand pose & 2 gloves & 60\,Hz & Per-finger articulation, both hands \\
ACE-Sense-Glove Lite & Contact pressure & 2 gloves & -- & Full-palm pressure map, both hands \\
% Microphone array \todo{model} & Spatial audio & 1 (??-ch) & ??\,kHz & Geometry calibrated \\
\midrule
Jetson Orin & Recording host & 1 & -- & Ingests all ZED One streams; NatNet-coordinated \\
Motive host (PC) & Mocap host, sync reference & 2 & -- & Renders the optical clock for cross-system sync \\
\bottomrule
\end{tabularx}
\end{table}

\subsubsection{Hardware Setup}
Within these environments, the sensor suite is assembled so that every facet of an interaction, from what the human sees, to how the body and objects move, to what the interaction sounds and feels like, has a dedicated sensor. Our two systems share this suite and differ only in sensor selection and placement, as summarized in Table~\ref{tab:hardware_inventory}.

\paragraph{Egocentric camera.} 
Participants in both systems wear an ACE-Ego-Head-V02 Lite by ACE Robotics, a head-mounted egocentric capture device with four fisheye cameras facing front-left, front-right, rear-left, and rear-right, recording $1088 \times 1280$ at $20$ FPS with an onboard IMU. 
Five OptiTrack markers are attached to the device to provide its initial placement and its real-time 6-DoF pose throughout the capture. This helps register the egocentric streams into the same world frame as every other sensor.

\paragraph{Exocentric camera.}
The table-scale configuration surrounds each workspace with $8$ close-range GoPro RGB cameras ($1920\times1080$ $@$ $30$ FPS), rigidly mounted on stands, while the room-scale configuration covers the apartment with $8$ wide-baseline ZED One RGB cameras ($1920\times1080$ $@$ $30$ FPS).

\paragraph{Human motion.}
Each participant wears a motion-capture suit with $41$ markers attached at the joints across the whole body (including $3$ markers around the head), tracked by the OptiTrack system (PrimeX 22; 12 cameras in the room-scale configuration, 16 in the table-scale configuration) at $60$ Hz to yield 41-joint skeletons, complemented by articulated hand poses, acquired via Manus motion-capture gloves at $60$ Hz in the room-scale configuration, and via RANSAC triangulation of 2D hand keypoints from the exocentric cameras followed by manual refinement in the table-scale configuration.

\paragraph{Object motion.}
Every interactable object is first digitized into a mesh by 3D scanning or 2D Gaussian Splatting reconstruction~\cite{huang20242d}, and fitted with OptiTrack markers that are bound to this mesh in the tracking system, so that the resulting 6-DoF trajectories at 60 Hz place the exact object geometry in the world frame.

\paragraph{Audio.}
We capture the audio via the GoPro exocentric cameras and the ACE-Ego-Head-V02 Lite to record the contact events, appliance operation, and ambient scene sound.

\paragraph{Tactile.}
Participants wear full-palm tactile gloves, which record contact pressure maps across the palm and fingers area.

\section{ACE-Data-0}
With the capture systems in place, this section presents ACE-Data-0. Specifically, Section~\ref{sec:ace_pipe} presents the data acquisition pipeline, covering the capture workflow, multi-modal recording, sensor synchronization, and calibration.
In Section~\ref{sec:ace_data}, we illustrate the data collection process, including task design, capture settings, dataset statistics, and modalities.
Finally, Section~\ref{sec:ace_annotation} introduces the annotations provided with the dataset and the pipeline used to produce them.

\begin{figure*}[!t]
  \centering
  \includegraphics[width=\linewidth]{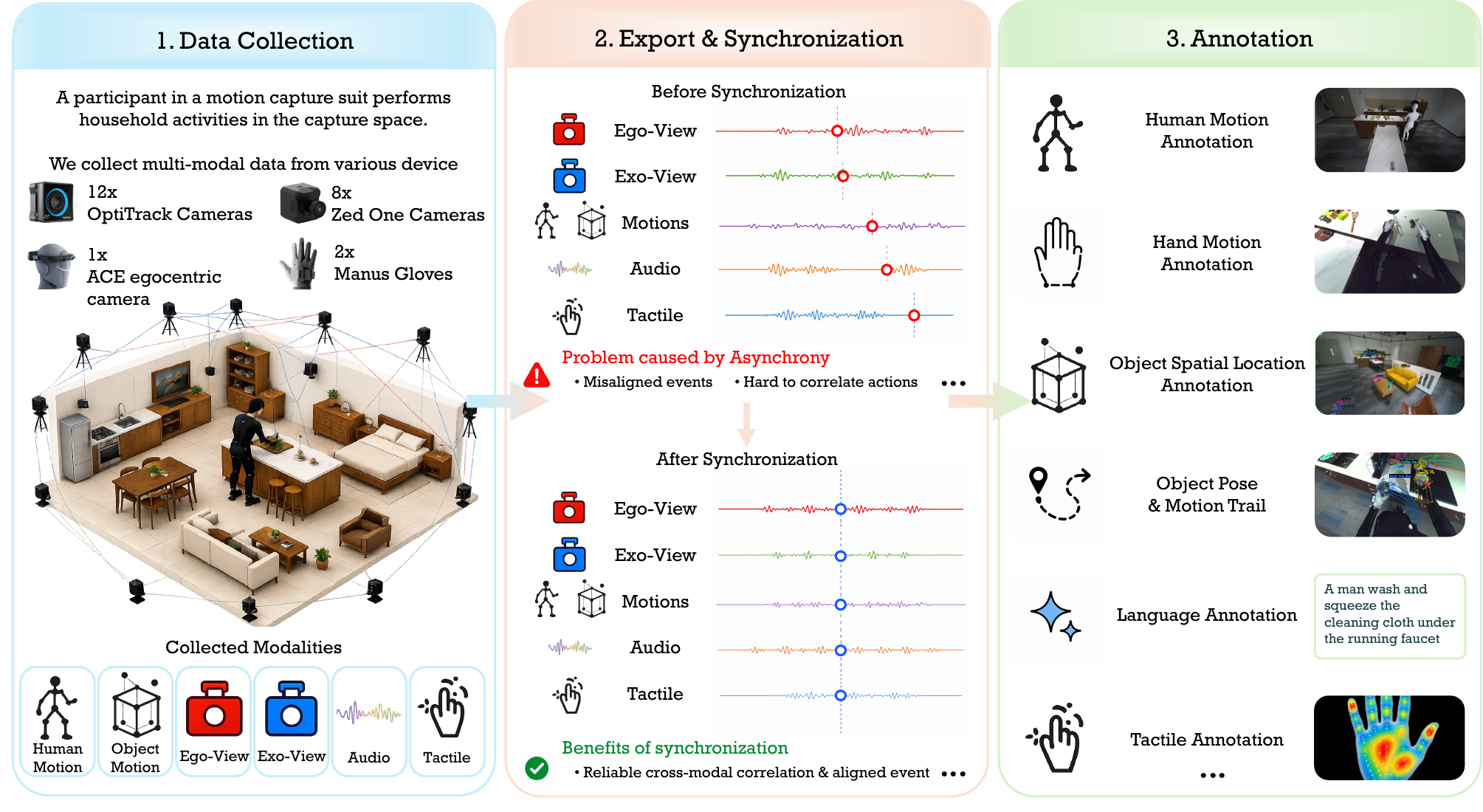}
  % \vspace{-4mm}
  \caption{\textbf{Overall workflow of ACE.} Each participant wears the motion-capture suit before the multi-modal recording. Synchronization is conducted after data export to align the timeline of each modality. Finally, we annotate the collected data with rich and high-quality labels derived from the captured ground truth.
  }
  \label{fig:workflow}
  % \vspace{-1em}
\end{figure*}
\subsection{Data Acquisition Pipeline}
\label{sec:ace_pipe}
The sensors above produce a dozen heterogeneous streams, and these streams are useful only when they can be precisely related to one another in both time and space. This subsection first describes how the streams are aligned, temporally (Section~\ref{subsec:synchronize}) and spatially (Section~\ref{subsec:calibration}), and then how capture sessions are operated in practice (Section~\ref{subsec:workflow} and Section~\ref{subsec:recording}). An overview of the workflow for ACE is presented in Fig.~\ref{fig:workflow}.

\subsubsection{Sensor Synchronization}
\label{subsec:synchronize}
Temporal alignment is the foundation of multi-modal capture: cameras, motion capture, object trackers, and tactile sensors run at different rates on independent clocks, and even a few milliseconds of drift are enough to corrupt contact-level annotation, making a hand appear to close on a cup several frames before the tactile stream registers the touch.
We take the OptiTrack clock as the reference, since the tracking system internally synchronizes its own cameras and delivers marker positions as strictly simultaneous $60$ Hz frames, and align each remaining device to it in turn.

\begin{figure*}[!t]
  \centering
  \includegraphics[width=\linewidth]{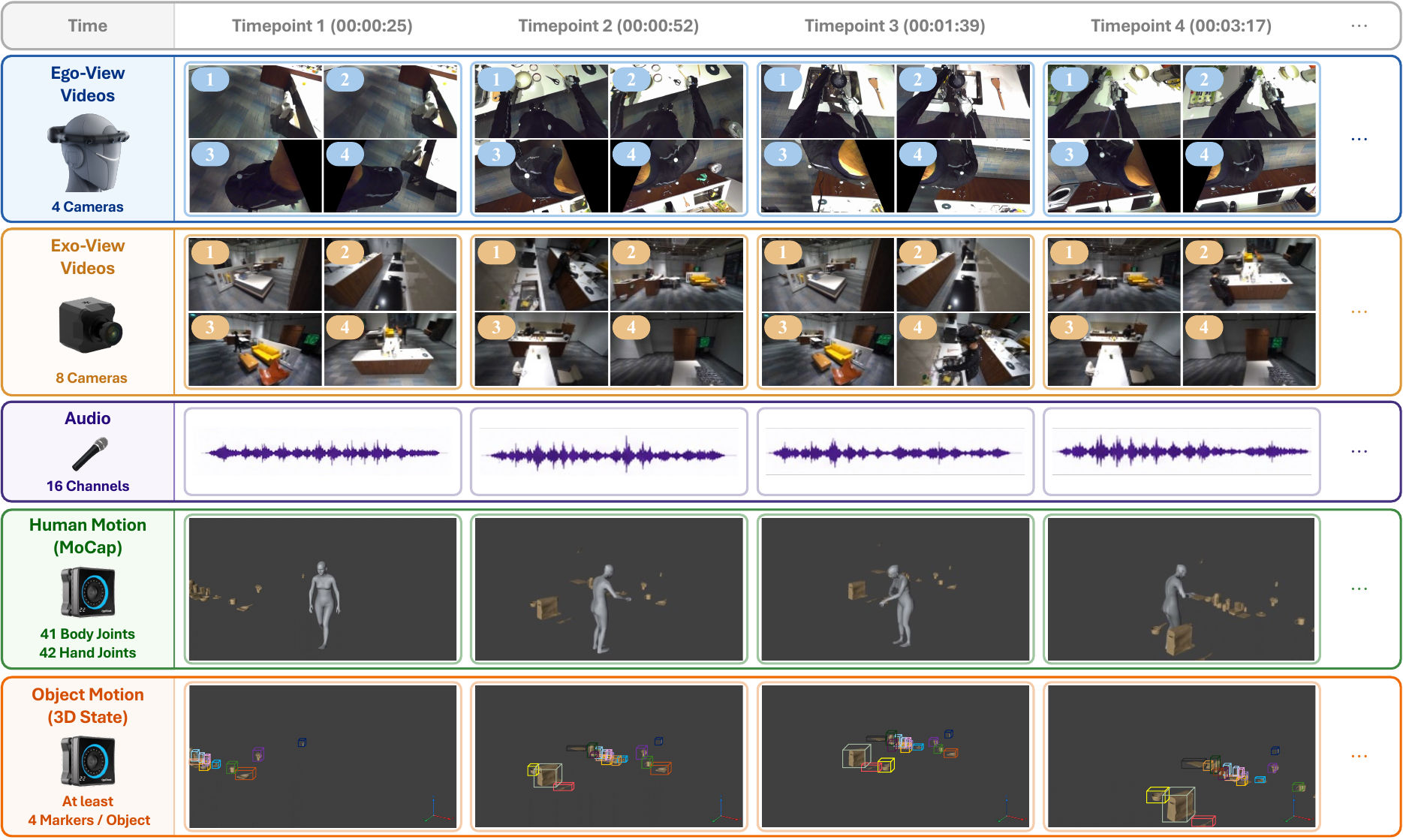}
  % \vspace{-4mm}
  \caption{\textbf{Synchronized modalities captured by ACE.} ACE is capable of capturing synchronized multi-modal data, including egocentric and exocentric videos, audio, and human and object motions, etc.
  }
  \label{fig:sync_data}
  % \vspace{-1em}
\end{figure*}
\paragraph{Exocentric cameras.}
(1) For the room-scale configuration, the ZED One cameras are ingested by a single NVIDIA Jetson Orin host over GMSL2, whose capture cards drive all cameras from a common frame trigger, so their captures are aligned by construction. 
Alignment to the OptiTrack clock then proceeds in two steps.
First, recording is coordinated over the local network: the camera host subscribes to the OptiTrack data stream via NatNet and starts and stops in lockstep with the motion-capture recording. 
This brackets the takes, but a network protocol without a shared clock cannot align individual frames. 
Unfortunately, the recorded timestamps cannot close this gap: they lag the true exposure moment by roughly $0.29$s, and this lag also drifts slowly over time.
To address this problem, we instead let the exocentric camera photograph a clock, in the form of QR codes. 
Specifically, the motion-capture host computer displays its own time on the monitor at nanosecond resolution.
Reading this clock off a recorded frame gives the exact capture time of that frame, in motion-capture time.
We sample such readings across a take and fit a line through them (a constant offset plus a slow drift). 
As a result, the final residuals after alignment are at the millisecond level, within a single motion-capture frame.
(2) For the table-scale configuration, the GoPro cameras are synchronized with one another by aligning their audio streams.

\paragraph{Egocentric cameras.}
The four fisheye views of ACE-Ego-Head are driven by the device's onboard controller, with a measured mutual misalignment of under $2$ ms.
(1) For the room-scale configuration, the alignment to the OptiTrack clock reuses the optical clock above, with one complication: the egocentric cameras face the hands and the wearer's back, and never see the monitor during a take.
Therefore, we start each take with a deliberate glance, in which the wearer points one camera of ACE-Ego-Head at the monitor for around $10$ seconds.
Reading the clock from these frames gives the offset between the clock of the egocentric camera and the OptiTrack clock.
The device's shared clock then carries this offset to the other three views for overall synchronization.
(2) For the table-scale configuration, the GoPro cameras and the egocentric headset read the same QR-code clock displayed on the monitor of the motion-capture host, as in the room-scale setup; this aligns the GoPro streams, already mutually synchronized via audio, to the headset clock.
The egocentric headset is then aligned to the OptiTrack system by registering its internally estimated poses (from AprilTag-based tracking) to the corresponding poses tracked via optical markers, completing the chain from the GoPro cameras, through the headset, to the OptiTrack timeline.
Together, these procedures register all devices in both configurations to the OptiTrack timeline.

\paragraph{Other sensors.}
The tactile gloves are synchronized with ACE-Ego-Head using their onboard IMU signals. 
At the beginning of each take, the operator performs a short motion pattern while keeping the hand and head approximately rigid, producing correlated IMU signals between the glove and the egocentric headset. 
We align the two streams by matching this motion template, establishing temporal synchronization.

\paragraph{Verification and output.}
As an independent check, we find that the per-camera time offsets re-estimated during calibration (Section~\ref{subsec:calibration}) are under $4$ ms, confirming the alignment.
All fitted offsets are folded into a per-take table that maps every camera frame, exocentric and egocentric alike, to its $60$ Hz motion-capture frame.
Downstream processing consumes only this table. See Fig.~\ref{fig:sync_data} for an example of synchronized capturing.

\begin{figure*}[!t]
  \centering
  \includegraphics[width=\linewidth]{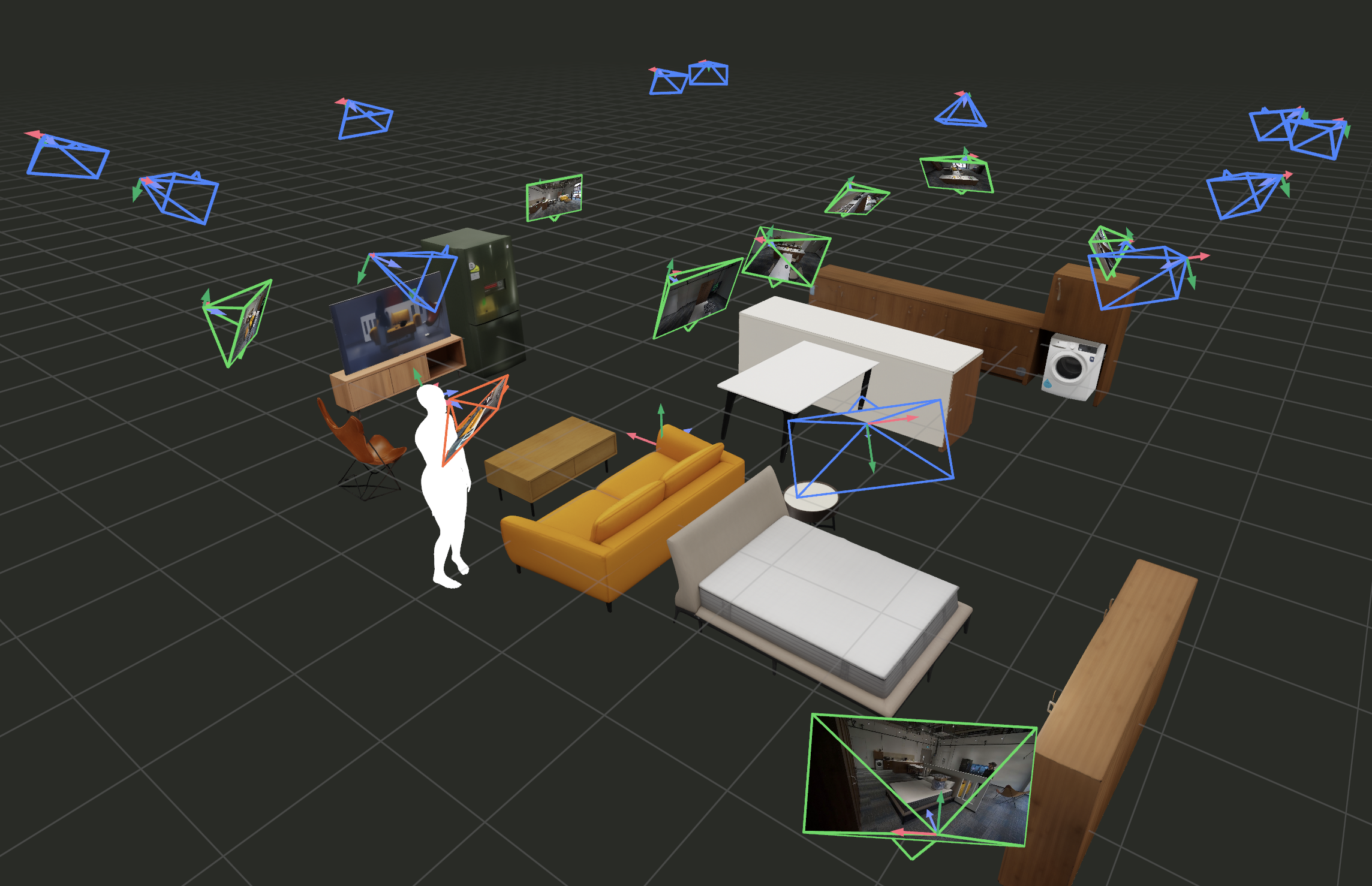}
  % \vspace{-4mm}
  \caption{\textbf{Calibrated cameras in the room-scale configuration.}
  The blue cameras represent the OptiTrack PrimeX 22 for motion capture; the green cameras denote the ZED One for capturing exocentric RGB videos; the orange camera is one of the egocentric cameras in ACE-Ego-Head.
  }
  \label{fig:calibration-sg}
  % \vspace{-1em}
\end{figure*}
\begin{figure*}[!t]
  \centering
  \includegraphics[width=\linewidth]{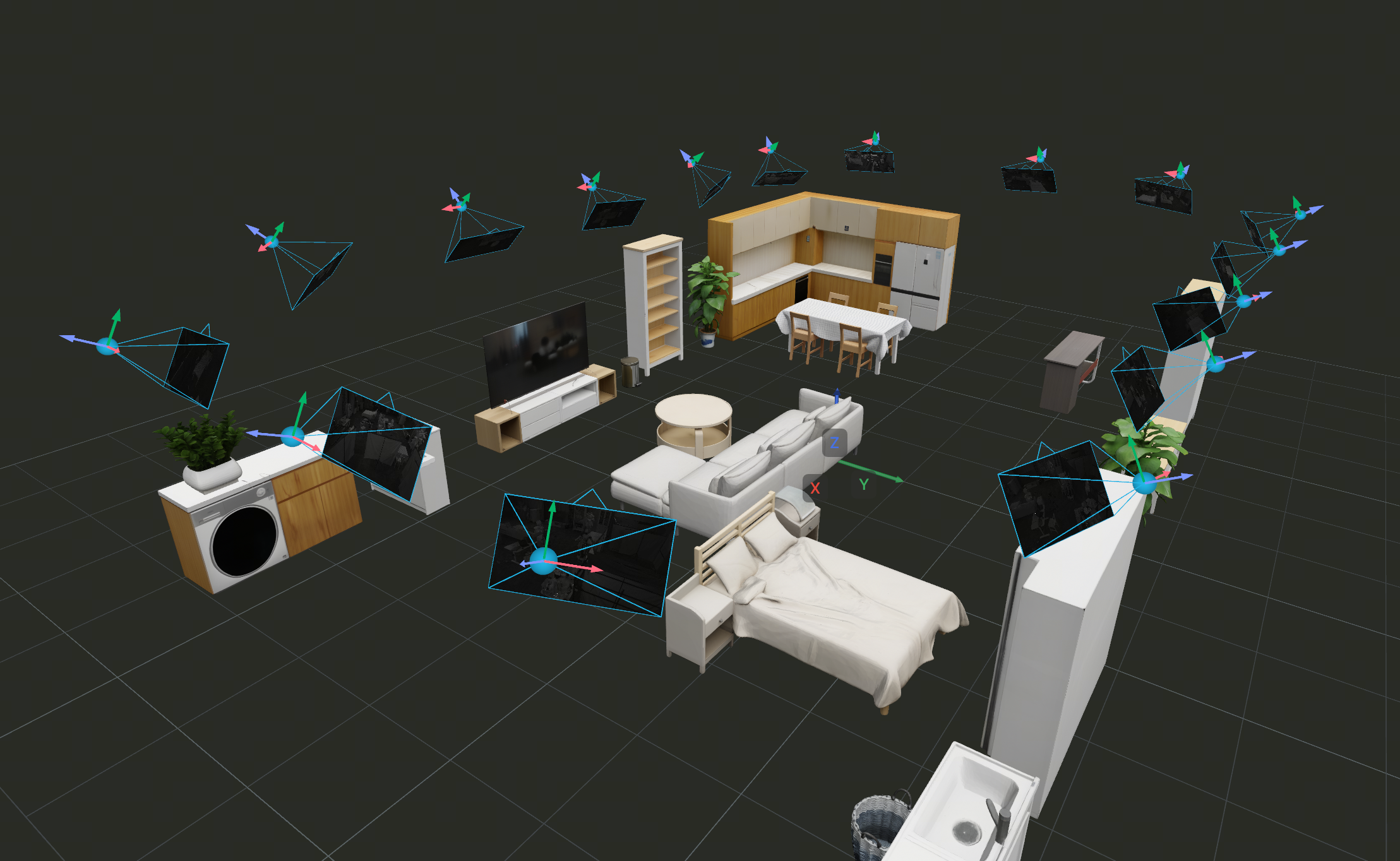}
  % \vspace{-4mm}
  \caption{\textbf{Calibrated cameras in the table-scale configuration.}
  The blue cameras represent the OptiTrack PrimeX 22 for motion capture.
  }
  \label{fig:calibration-sz}
  % \vspace{-1em}
\end{figure*}
\subsubsection{Calibration}
\label{subsec:calibration}
Whereas the synchronization aligns the streams in time, calibration registers them in space, and the two camera systems pose opposite challenges: the exocentric cameras never move but barely share a view, while the egocentric cameras move every frame. The two calibration procedures are illustrated in Fig.~\ref{fig:calibration-sg} and Fig.~\ref{fig:calibration-sz}.

\paragraph{Exocentric cameras.}
Standard multi-camera calibration assumes co-visibility: two cameras must observe the same target at the same time. 
Our sparse placement of the exocentric cameras breaks this assumption, as most camera pairs share no common area. 
We therefore bridge the cameras through the motion-capture system instead.
Specifically, we utilize an ArUco board~\cite{GARRIDOJURADO20142280} with a retroreflective marker at each corner as the calibration target.
This one physical object is visible to every sensing system at once: the RGB cameras see the printed pattern of the ArUco board, while the OptiTrack infrared cameras see both that pattern and the retroreflective corner markers. The board therefore ties the RGB cameras and the motion-capture volume to a single reference without requiring any two cameras to share a view.
Then, the calibration proceeds in three steps:
\textbf{(1)} First, the infrared cameras estimate where the corner markers actually sit on the board. It is important to note that these markers are attached by hand, so their exact positions are not known in advance.
Aggregating the estimates over thousands of board poses reduces the error in this step to the millimeter level.
\textbf{(2)} Second, with their positions known, the markers alone give the board's pose in every frame, at sub-millimeter consistency.
\textbf{(3)} Third, each RGB camera is fitted against this marker-anchored board trajectory. Not every observation contributes in this step: (i) frames in which the board was moving are discarded, because residual timing error grows with the motion, and (ii) the detections near the image edges are discarded, where the distortion model is least reliable.
A final joint refinement then re-estimates the board pose from all cameras together and refits each camera in turn. Throughout, the tracked markers remain the dominant reference, keeping the solution anchored in the motion-capture world. 
On held-out frames, every camera reaches a median reprojection error below $3$ px, roughly a centimeter of 3D error at typical distances.

\paragraph{Egocentric cameras.}
A moving rig is not described by one pose but by a pose for every frame.
Considering our situation where the egocentric views are dominated by the wearer's own arms and torso and the front and back camera pairs share no field of view, recovering this trajectory from vision alone, as SLAM would, is unreliable.
We therefore do not estimate where the cameras are; we measure them.
The five markers on the ACE-Ego-Head chassis form a rigid body that OptiTrack tracks at $60$ Hz. 
The only unknown left is the fixed transformation from each fisheye camera to this body, which is a classical hand-eye calibration problem~\cite{tsai1989new}.
We first solve it independently for each camera.
A joint bundle adjustment~\cite{triggs2000bundle} then refines the four transformations together, treating the board pose, shared by all cameras, and the per-camera time offsets as additional free variables.
We also use only low-angular-velocity frames for the fitting, as the timing error grows with the motion.
The final median reprojection error is about $2$ px.
Once calibrated, producing camera poses for a take requires no images at all.
OptiTrack tracks the egocentric camera's rig body at $60$ Hz, while the cameras record at $20$ FPS; for each egocentric frame, we interpolate the rig's tracked pose to the frame's timestamp and apply that camera's hand-eye transformation. 
Every pose is thus measured rather than estimated, and does not drift.

\paragraph{Other calibrations.}
The two procedures above answer where the cameras are.
The remaining calibrations describe each sensor itself.
For the cameras, this means intrinsics: the exocentric ones use their factory parameters, and the egocentric fisheyes are calibrated with Kalibr~\cite{Fur2013kalibr}.
For the objects, this means geometry: each marker set is registered once to its scanned or 2DGS-reconstructed mesh, so that tracking the markers places the full object in the world frame.
All of these are re-verified twice a day. Together with the synchronization above, we complete the unified spatio-temporal frame on which all subsequent annotation and benchmarking rest.

\subsubsection{Capture Workflow}
\label{subsec:workflow}
With alignment in place, every capture session follows the same five-step protocol at both sites.
\textbf{(1)} Scene preparation: objects are placed at randomized yet plausible initial positions.
\textbf{(2)} Participant setup: the participant puts on the motion-capture suit, the ACE-Ego-Head headset, and the gloves, then performs a short T-pose routine that registers their skeleton with the tracking system.
\textbf{(3)} Task briefing: the participant receives a goal-level instruction verbally; how to achieve the goal is left entirely to them.
\textbf{(4)} Recording: the take opens with the clock glance described in Section~\ref{subsec:synchronize}, after which all sensors record continuously while an operator monitors stream health on a live dashboard.
\textbf{(5)} Post-checks: synchronization and tracking quality are verified after each take, and failed takes are flagged for re-capture.

\subsubsection{Multi-modal Recording}
\label{subsec:recording}
During step (4) of the capture workflow, each system simultaneously acquires: (i) the exocentric RGB streams;
(ii) $4$ egocentric fisheye streams with IMU;
(iii) full-body motion and hand poses at $60$ Hz;
(iv) $6$-DoF poses of all tracked objects at $60$ Hz;
and (v) tactile signals.
All streams are timestamped against the common clock established in Section~\ref{subsec:synchronize}. A one-hour session produces approximately $1$ TB of raw data.

\subsection{Data Collection}
\label{sec:ace_data}
The preceding subsections established the recording machinery; the question that remains is what to record with it. We describe the task design (Section~\ref{sec:task}), the capture settings (Section~\ref{sec:setting}), the resulting dataset statistics (Section~\ref{sec:statistics}), and the modalities included in each released sequence (Section~\ref{sec:modality}).

\subsubsection{Task Design}
\label{sec:task}
The long-horizon character of ACE-Data-0 originates in its task design. By a \textit{long-horizon} task we mean an activity that unfolds over minutes rather than seconds and decomposes into a sequence of sub-actions whose order is constrained by the goal rather than fixed in advance.
For example, consider the task of ``prepare a cup of tea and serve it at the table''. It involves a chain of human-object interactions: the participant walks to the cupboard, opens it, takes out a cup, fills the kettle, waits for the water to boil, retrieves a tea bag, pours, stirs, carries the cup across the room, and clears space before setting it down. This single take may thus contain planning, locomotion, and fine-grained manipulations.

However, most HOI datasets instead prescribe atomic actions~\cite{brahmbhatt2020contactpose,chao2021dexycb,kwon2021h2o,fan2023arctic}, capturing a single grasp or handover in isolation. To address this gap, we prescribe household goals and let the actions emerge: how to reach the goal is left to the participant. Different participants order the sub-tasks differently, grasp differently, and reach for different objects, so the variability of real behavior enters the data by itself.

Specifically, we record three types of takes:
\begin{itemize}
    \item \textbf{Atomic HOI tasks} contain one to three household tasks each, drawn from more than 15 types of household activities, such as pouring water, drinking, making tea, watering plants, chopping vegetables, cooking, and tidying up; each take lasts roughly three minutes. Each of them provides clean, self-contained instances of everyday manipulation, the unit from which skills are most readily learned.
    \item \textbf{Chains of HOI tasks} combine the full range of short tasks into one continuous activity of roughly twenty to thirty minutes. Sub-tasks interleave freely, yet every take ends with the scene tidied back into order, tracing a full cycle of household activity. Compared to the atomic HOI tasks, they exercise long-horizon planning, state tracking, and memory at a further level of complexity.
    \item \textbf{HSI tasks} involve almost no objects, focusing instead on interactions with scene components such as tables, chairs, and sofas. Such takes record whole-body motion, such as walking and exercising, and human-scene contact, such as sitting, lying, and leaning, in about five minutes per take. These movements complete the range of behaviors a humanoid must master.
\end{itemize}

\begin{figure*}[!t]
  \centering
  \includegraphics[width=\linewidth]{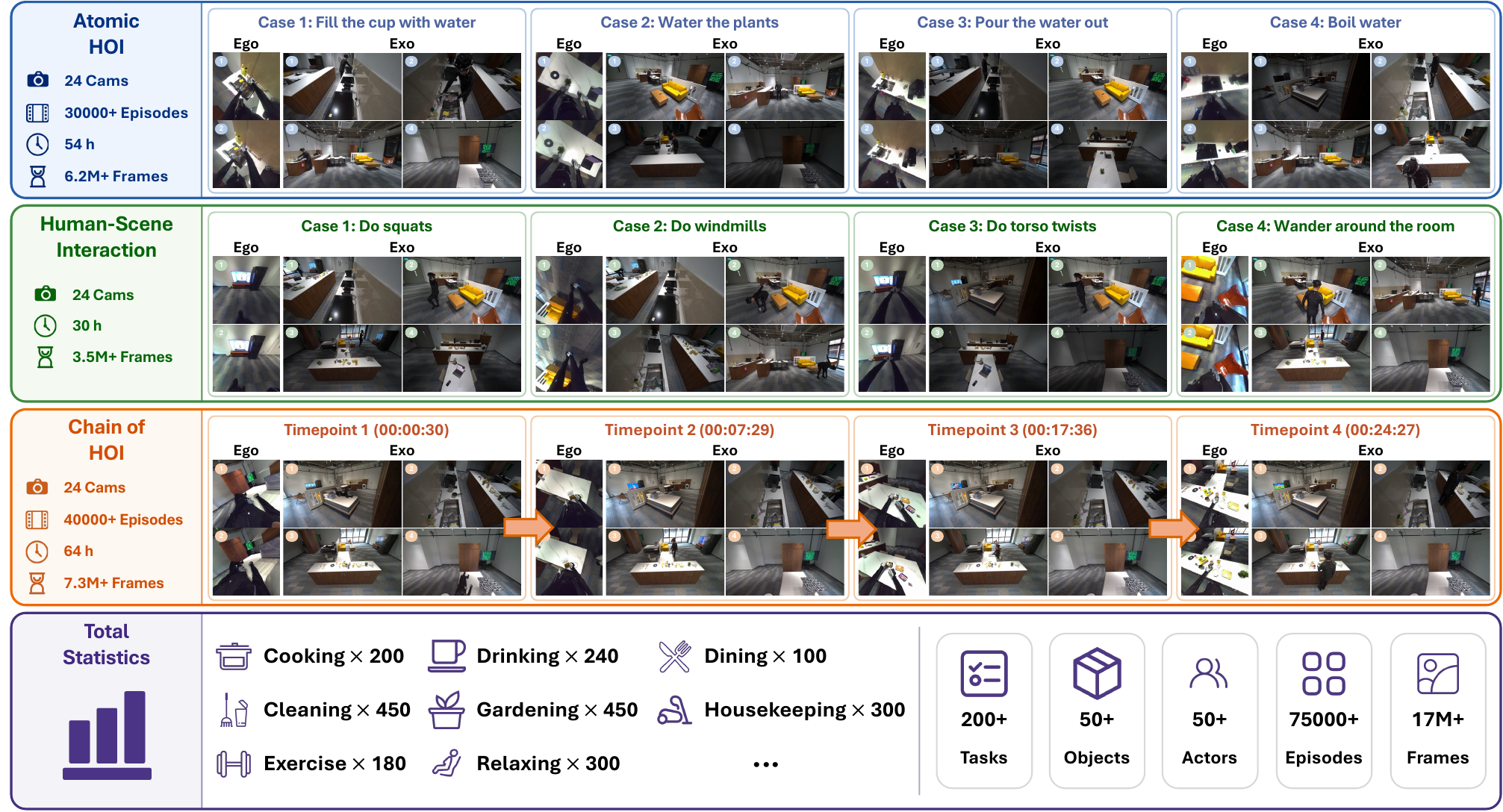}
  % \vspace{-4mm}
  \caption{\textbf{Examples of different task categories captured in ACE-Data-0.}
  }
  \label{fig:script_example_sg}
  % \vspace{-1em}
\end{figure*}
To decide what these recordings should contain, we surveyed the tasks that arise in everyday home life and designed the atomic HOI, chain-of-HOI, and HSI tasks accordingly. We present examples of different task categories in Fig.~\ref{fig:script_example_sg}.

\begin{figure*}[!t]
  \centering
  \includegraphics[width=\linewidth]{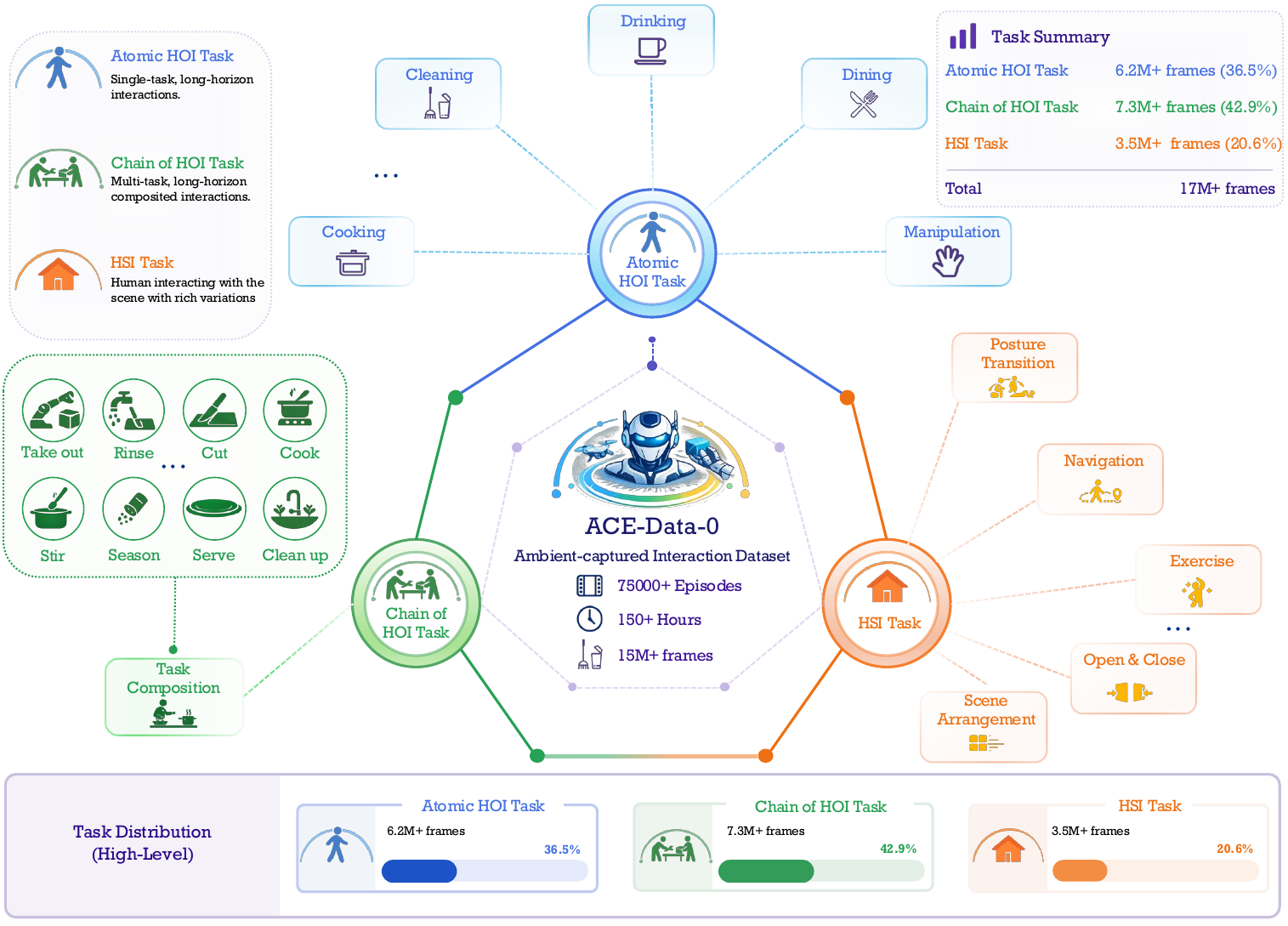}
  % \vspace{-4mm}
  \caption{\textbf{Overview statistics of ACE-Data-0.} ACE-Data-0 covers three types of interaction tasks: atomic HOI tasks, chains of HOI tasks, and HSI tasks.
  }
  \label{fig:statistic}
  % \vspace{-1em}
\end{figure*}

\subsubsection{Capture Settings}
\label{sec:setting}
To perform these tasks, we recruited 50 participants.
Each participant completes all the designed tasks across a $2$-day session.
Diversity enters the data from two directions.
On the human side, the goal-level instructions leave the behavior open: participants differ in where they start and end, the route they take in between, the interactions they choose, and how long each step takes. 
On the object side, takes vary in which categories appear and in what number, where the objects are placed, where they begin and end, and along what trajectories and in what manner they are moved.
% All participants signed informed consent under a protocol approved by the ethics board, which covers the public release of the recordings, including faces.

\subsubsection{Dataset Statistics}
\label{sec:statistics}
This collection effort yields more than 150 hours of synchronized multi-modal capture, comprising over 17M frames across more than 75,000 episodes. We define an \textit{episode} as a contiguous segment of interaction that realizes one meaningful sub-goal, the smallest unit that remains semantically self-contained when used as a training example. Episodes are counted within takes rather than recorded in isolation, so the surrounding context, namely the actions that precede and follow each segment, is preserved in the same stream. In ACE-Data-0, even the shortest takes run minutes rather than seconds, roughly an order of magnitude longer than typical HOI clips~\cite{brahmbhatt2020contactpose,chao2021dexycb,kwon2021h2o,fan2023arctic}. Fig.~\ref{fig:statistic} reports the distributions of task categories, take lengths, object categories, etc.

\subsubsection{Modalities}
\label{sec:modality}
A released take contains more than the raw streams of Section~\ref{subsec:recording}: it also carries everything the pipeline of Section~\ref{sec:ace_pipe} derives from them, in a common frame and on a common timeline. Concretely, each take bundles:
\begin{itemize}
    \item \textbf{Egocentric video}: the four fisheye views, their IMU readings, and per-frame 6-DoF headset poses from the tracked rig;
    \item \textbf{Exocentric video}: all views, each with intrinsics and its pose in the world frame;
    \item \textbf{Human motion}: $41$-joint body skeletons with articulated hand poses and converted SMPL-X motion parameters~\cite{pavlakos2019expressivebodycapture3d};
    \item \textbf{Object motion}: per-object 6-DoF trajectories, with scanned or 2DGS-reconstructed meshes for more than 50 instances;
    \item \textbf{Audio}: synchronized multi-source audio recorded from GoPro exocentric cameras and the egocentric headset;
    \item \textbf{Tactile}: hand-shaped pressure grids remapped from raw glove sensors, with calibrated normalization and baseline correction.
\end{itemize}

\subsection{Annotation}
\label{sec:ace_annotation}

The raw signals described so far become substantially more useful once enriched with annotations. We describe what annotations ACE-Data-0 provides (Section~\ref{sec:anno_types}) and how they are produced at scale (Section~\ref{sec:anno_pipeline}).

\begin{figure*}[!t]
  \centering
  \includegraphics[width=\linewidth]{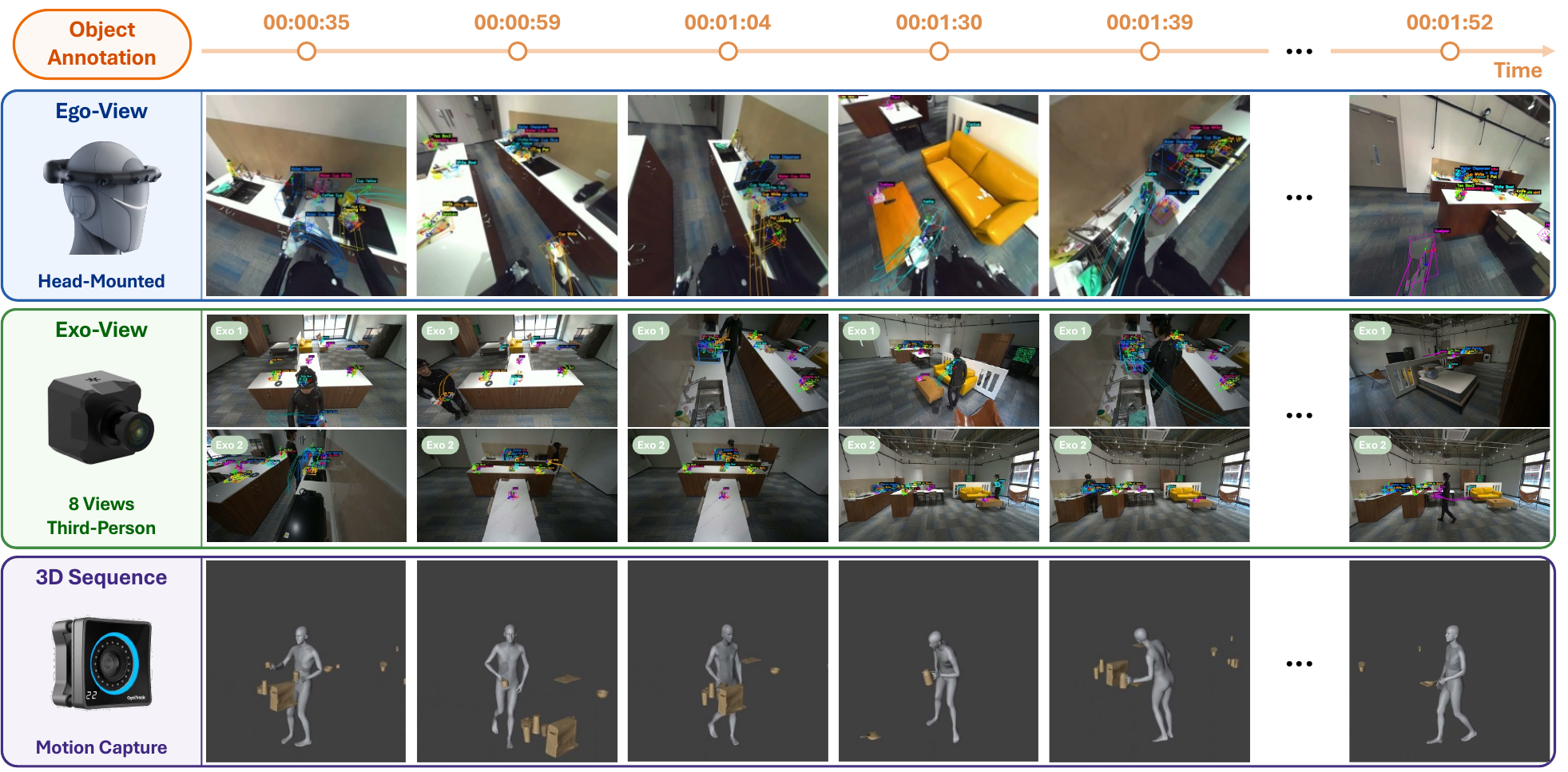}
  % \vspace{-4mm}
  \caption{\textbf{Examples of data annotation for captured objects.}
  ACE-Data-0 provides rich annotations for the captured objects, including name tags, 6-DoF poses, bounding boxes, and motion trails.
  }
  \label{fig:annotation_obj_sg}
  % \vspace{-1em}
\end{figure*}
\begin{figure*}[!t]
  \centering
  \includegraphics[width=\linewidth]{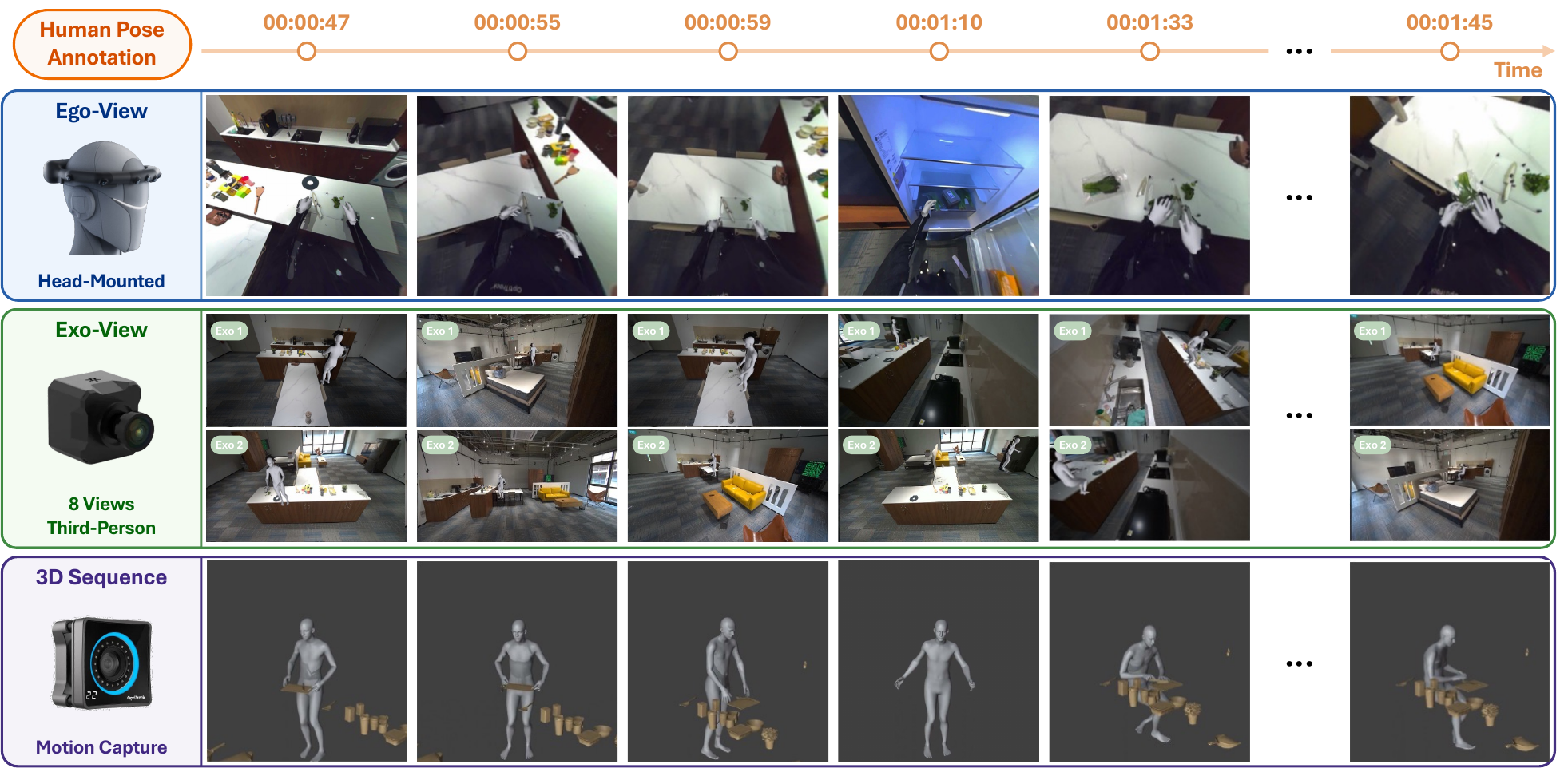}
  % \vspace{-4mm}
  \caption{\textbf{Examples of data annotation for the captured human body and hands.}
  ACE-Data-0 provides ground-truth poses for both the human body and hands. We also project the poses to both egocentric and exocentric videos for better embodied AI learning with visual inputs.
  }
  \label{fig:annotation_human_sg}
  % \vspace{-1em}
\end{figure*}
\begin{figure*}[!t]
  \centering
  \includegraphics[width=\linewidth]{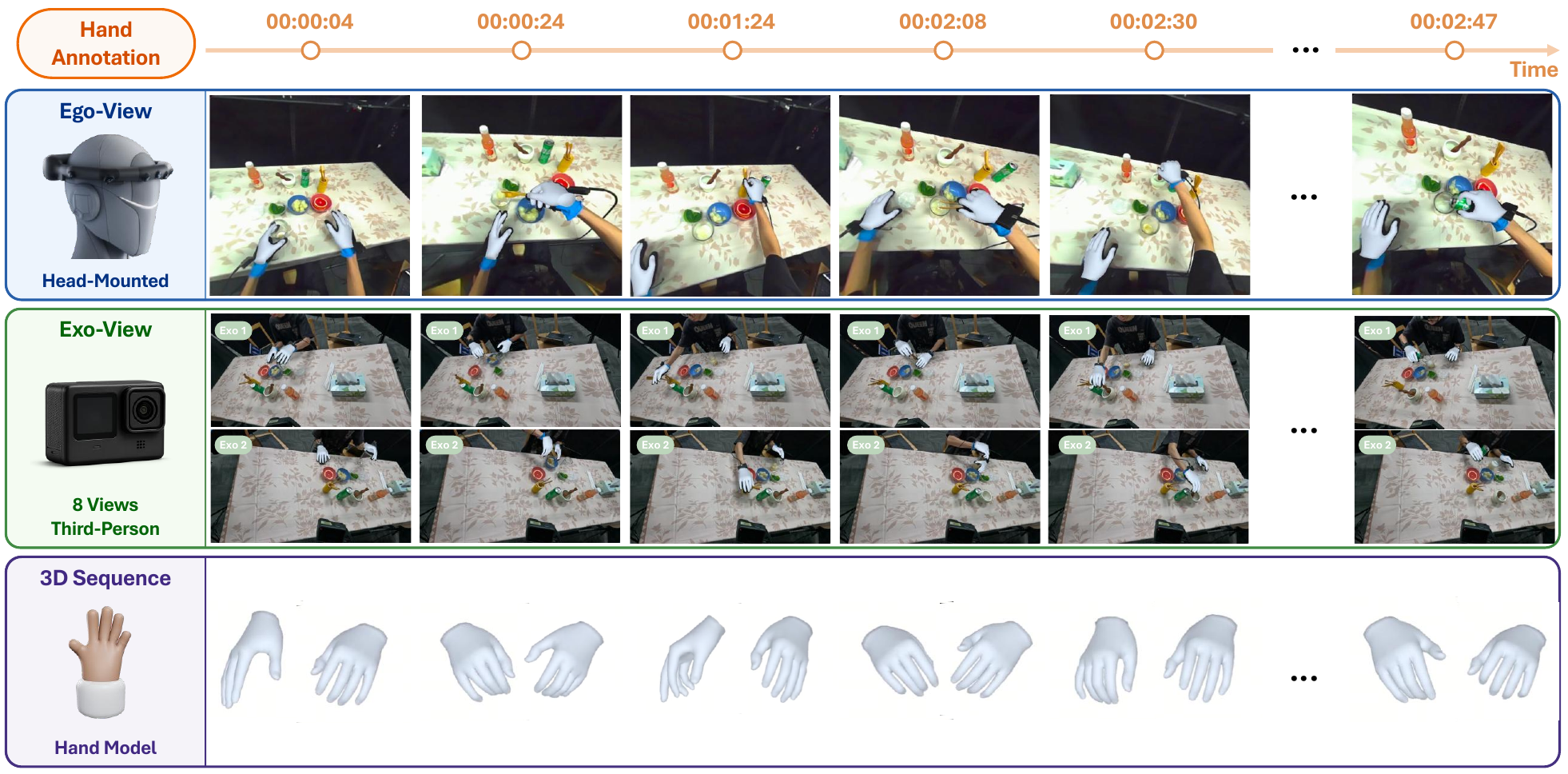}
  % \vspace{-4mm}
  \caption{\textbf{Examples of data annotation for captured hands.}
  ACE-Data-0 provides ground-truth poses for human hands. We also project the poses to both egocentric and exocentric videos for better embodied AI learning with visual inputs.
  }
  \label{fig:annotation_hand}
  % \vspace{-1em}
\end{figure*}
\begin{figure*}[!t]
  \centering
  \includegraphics[width=\linewidth]{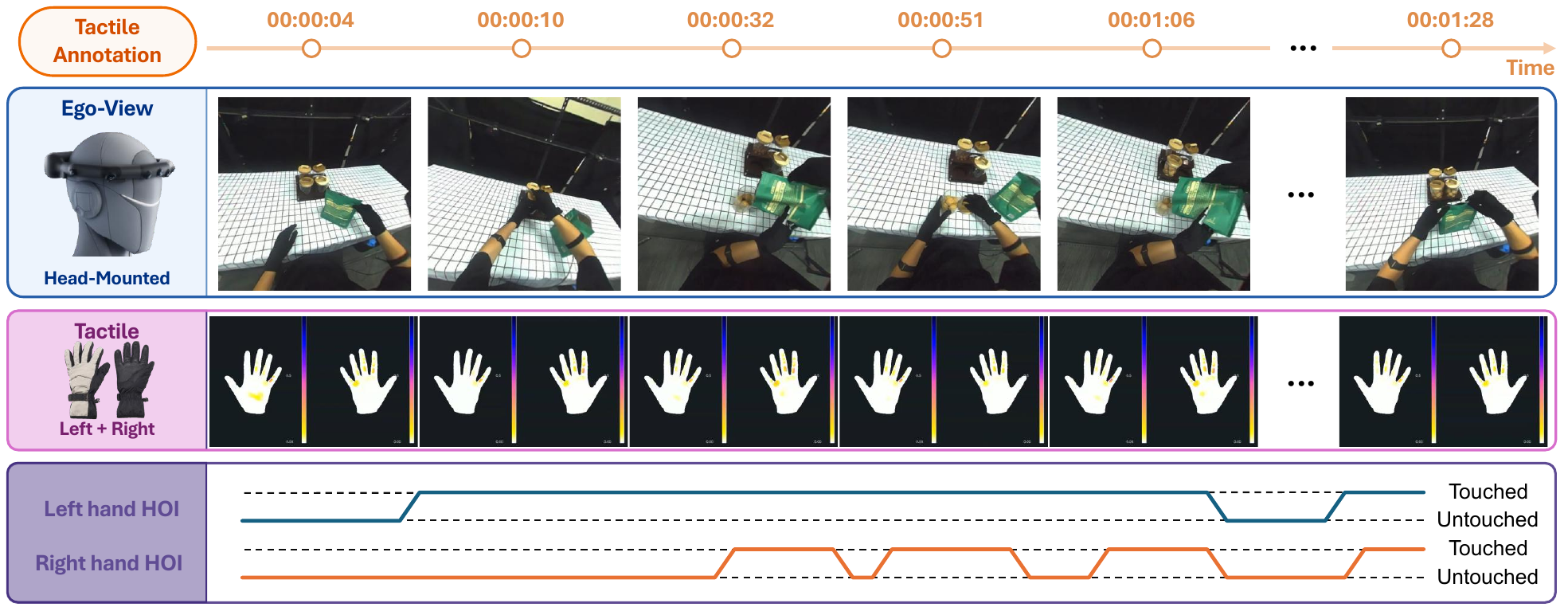}
  % \vspace{-4mm}
  \caption{\textbf{Examples of data annotation for tactile signals.}
  ACE-Data-0 provides ground-truth full-hand grasp pressure values, resolving interaction events that remain ambiguous under visual occlusion.
  }
  \label{fig:annotation_tactile}
  % \vspace{-1em}
\end{figure*}
\begin{figure*}[!t]
  \centering
  \includegraphics[width=\linewidth]{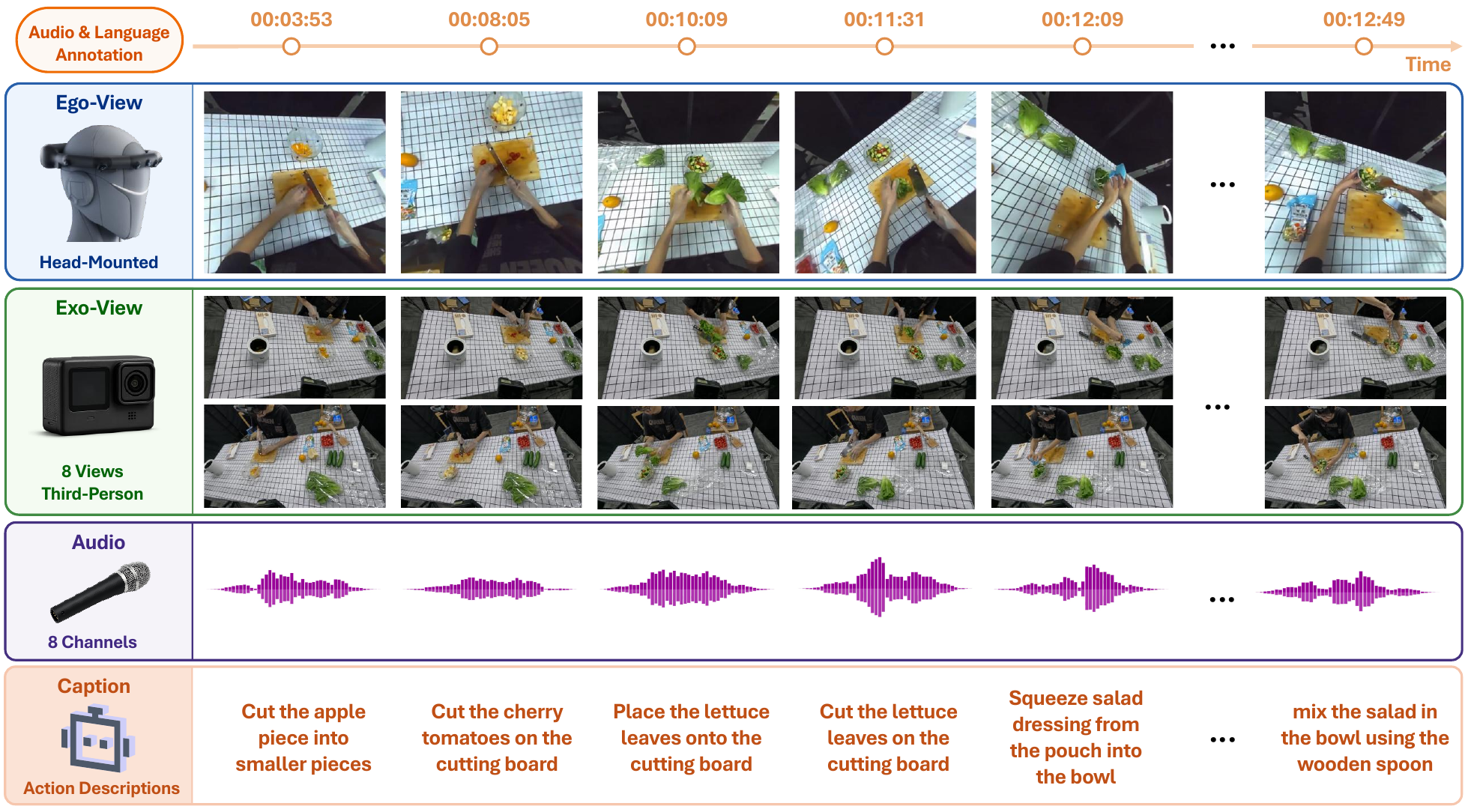}
  % \vspace{-4mm}
  \caption{\textbf{Examples of data annotation for audio and language description.}
  ACE-Data-0 provides natural language descriptions aligned with the timeline of measured activities.
  }
  \label{fig:annotation_caption}
  % \vspace{-1em}
\end{figure*}
 
\subsubsection{Annotation Types}
\label{sec:anno_types}
ACE-Data-0 provides five types of annotation, which jointly characterize every take: the objects involved in the interaction, the configuration of the actor's body, the articulation of the hands during manipulation, the physical contact established between hand and object, and the acoustic and semantic description of the activity itself.
Specifically, the object annotations (Fig.~\ref{fig:annotation_obj_sg}) comprise a category label, a bounding box, and a per-frame 6-DoF pose for every tracked object, together with the motion trail traced by the object over the course of the take. 
Human annotations (Fig.~\ref{fig:annotation_human_sg}) provide full body pose throughout each take, and hand annotations (Fig.~\ref{fig:annotation_hand}) refine this to articulated finger configurations during periods of dexterous manipulation. 
Tactile annotations (Fig.~\ref{fig:annotation_tactile}) register the timing and spatial distribution of contact, resolving interaction events that remain ambiguous under visual occlusion. 
Audio and language annotations (Fig.~\ref{fig:annotation_caption}) pair the recorded audio streams with natural-language descriptions that identify the sound events produced by the interaction, such as an object set down on a surface or a container being opened, and state the goal of each take alongside the sequence of sub-goals through which it is achieved.
 
\paragraph{Calibration and synchronization.} Every take ships with per-camera intrinsics, camera poses in the shared world frame, and the timeline that aligns all sensor streams. These are the outputs of Section~\ref{sec:ace_pipe}, released as data: with them, any tracked 3D point can be projected onto any pixel of any view, and any two streams can be paired at any instant.  Users can therefore combine the modalities freely for training, pairing any subset of them as inputs and supervision, without rerunning any part of our pipeline.

\paragraph{Human poses.} The captured body and hand poses are reprojected onto every frame of every camera, giving per-view 2D mesh overlays consistent with the 3D capture. Because these projections come from measured 3D states rather than image-based detectors, they remain correct where detectors typically fail: under furniture occlusion, extreme viewpoints, and motion blur. Each frame thus carries pixel-aligned 2D poses in every view and metric 3D poses in the world frame.
 
\paragraph{Tactile labels.} Each frame carries the tactile reading from the tactile glove, temporally aligned with the visual streams and the object label currently in use. Contact thus can be detected rather than inferred: the signal comes directly from the sensor surface, not from appearance or from proximity between reconstructed geometry. These readings mark when and with what an interaction happens, the anchor from which most HOI tasks start.
 
\paragraph{Object annotations.} Every object instance carries its scanned or 2DGS-reconstructed mesh, per-frame 6-DoF poses, 2D/3D bounding boxes in all views, and its motion trail over the take. Mesh and pose together place the exact object geometry in the scene at every instant; the boxes and trails are their projections into each camera. The full history of an object, where it sat, when it moved, and where it ended, can therefore be queried at any point of a take.

\paragraph{Audio descriptions.}
Each take carries multi-channel audio recorded by the head-mounted egocentric capture rig and the GoPro cameras, sharing the synchronization clock of the visual streams and therefore aligned to the same timeline as the pose, object, and tactile annotations. Two classes of sound are present: those produced by the interaction itself, such as an object set down on a surface or liquid poured into a cup, and the ambient acoustics of the domestic environment, including appliance noise, footsteps, and room reverberation. Because the audio is captured from the actor's own viewpoint, the acoustic perspective moves with the participant, so a sound event varies in loudness and spatial character with proximity.

\paragraph{Textual descriptions.} Gemini-3.1-pro-preview~\cite{geminiteam2025geminifamilyhighlycapable} watches the ego-view video and describes each time span in natural language: what the person is doing, and what happens in the scene. The descriptions give each take a searchable storyline and connect the physical record to language, in the form that vision-language and vision-language-action models consume directly. 

\subsubsection{Annotation Pipeline}
\label{sec:anno_pipeline}
Producing annotations of this breadth would ordinarily demand prohibitive manual effort.
ACE avoids most of it, because among our five annotation types, all but the textual descriptions are \emph{measured} rather than estimated.
Human and object states are metrically tracked, every camera is calibrated, and contact is directly sensed. 
The pose reprojections, bounding boxes, motion trails, and contact events therefore follow from the recorded states by projection and tactile sensing, with no estimation model in the loop. 
The textual descriptions are the one generated type: Gemini produces them from the ego-view videos, segment by segment. 
Finally, human annotators check the auto-generated labels and correct the descriptions by hand.
\section{Benchmark}
\label{sec:benchmark}

\begin{figure*}[!t]
  \centering
  \includegraphics[width=\linewidth]{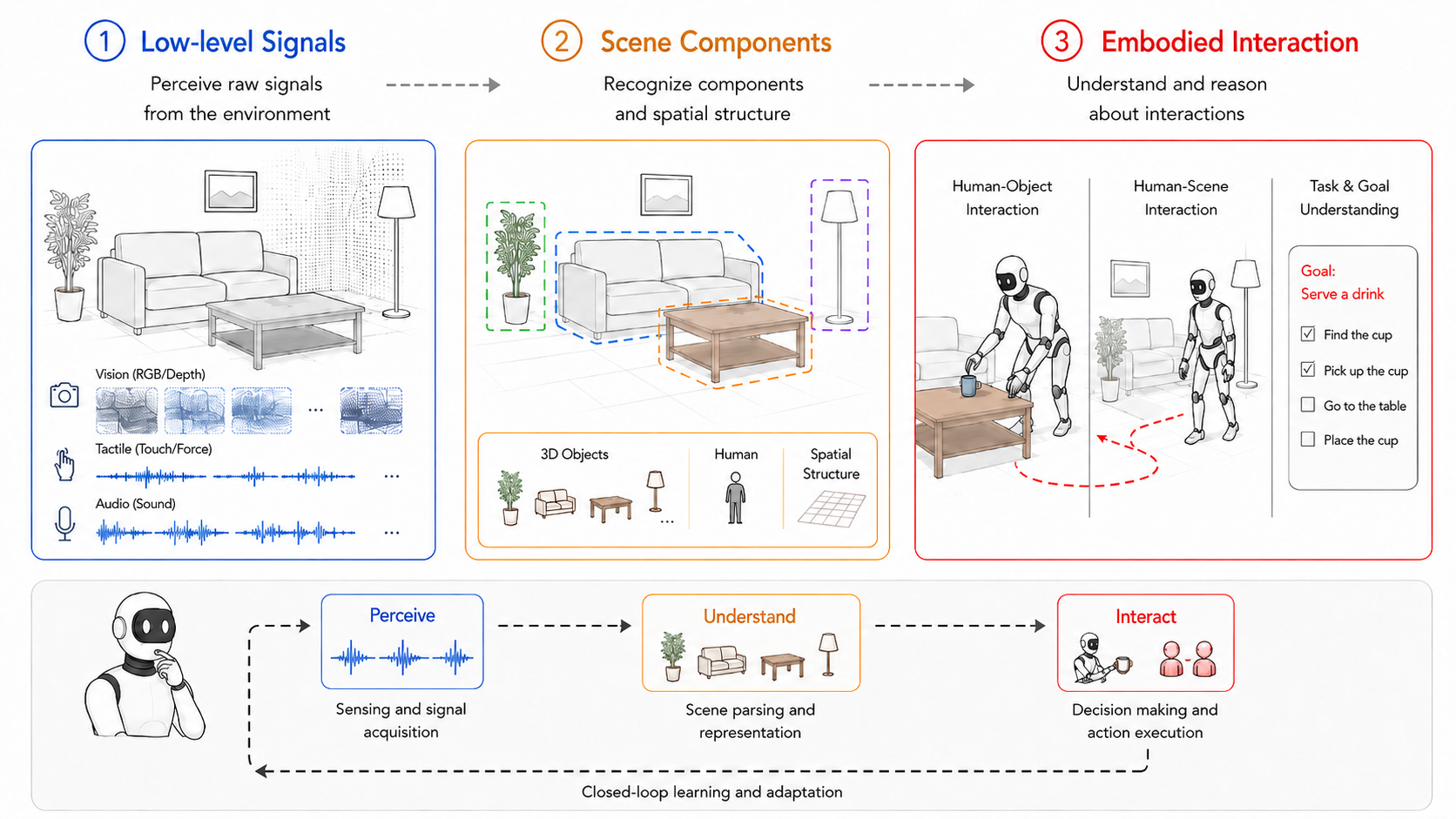}
  % \vspace{-4mm}
  \caption{\textbf{Three hierarchical benchmark levels.}
  Our benchmarked levels, \textit{i.e.}, low-level signal inference, scene component recovery, and interaction estimation, exactly mimic how human beings and embodied AI would perceive real-world environments.
  }
  \label{fig:benchmark_levels}
  % \vspace{-1em}
\end{figure*}
Having described how ACE-Data-0 is captured and annotated, we now use it to evaluate existing methods, with a diagnostic goal: to expose where current approaches break on long-horizon home-scene data, and to indicate where solutions might lie. 
The benchmark comprises three levels, advancing from \emph{signals}, to \emph{components}, to \emph{interactions}, with each level building upon the one beneath it: 
\emph{low-level signal inference} operates directly on the raw sensory streams, \textit{i.e.}, predicting tactile signals from video (Section~\ref{sec:bench_low}); 
\emph{scene component recovery} assembles these signals into the 3D human components (Section~\ref{sec:bench_scene}); 
and \emph{interaction estimation} recovers the hand motion through which the human engages objects, from egocentric and exocentric views (Section~\ref{sec:bench_interaction}). 
This progression mirrors the perceptual capabilities an embodied agent must chain together when acting in a home: sensing contact, estimating scene state, and mastering hand-object coordination (see Fig.~\ref{fig:benchmark_levels}). 
We hold out 10 hours of capture as a test set for all benchmark evaluations in this report.
Unless otherwise noted, baselines are evaluated with their officially released pre-trained checkpoints, for fair comparison.

\subsection{Low-level Signals}
\label{sec:bench_low}
We begin with the raw sensory signals that interactions produce, and among them, touch holds a special position.
Vision tells an observer where things are; touch tells the actor whether a grasp has succeeded, how firmly to hold, and when an object begins to slip. 
Robots need this signal as much as humans do, yet contact sensing remains scarce in practice: tactile hardware is expensive and fragile, and absent from most deployed platforms, while cameras are everywhere.
Learning to read touch out of video would therefore turn the most abundant sensor into a substitute for the scarcest one.
ACE-Data-0 makes this mapping learnable: through the synchronization of Section~\ref{subsec:synchronize}, every video frame is paired with the tactile reading recorded at the same instant.

\begin{table}[t]
    \centering
    \caption{\textbf{Tactile estimation from ego-view on ACE-Data-0}.}
    \label{tab:tactile}
    \footnotesize
    \setlength{\tabcolsep}{6pt}
    \begin{tabular*}{\linewidth}{@{\extracolsep{\fill}}l c c c c}
        \toprule
        \textbf{Method} & \textbf{Temp Acc.} $\uparrow$ & \textbf{C-IoU} $\uparrow$ & \textbf{V-IoU} $\uparrow$ & \textbf{CoP} $\downarrow$ \\
        \midrule
        PressureVision~\cite{DBLP:conf/eccv/GradyTBTWHK22} & 0.0093 & 0.0007 & 0.0000 & 10.9807 \\
        EgoPressureDiff~\cite{DBLP:conf/icml/abs-2606-09243} & 0.2912 & 0.0197 & 0.0025 & 8.5152 \\
        TouchAnything~\cite{DBLP:journals/corr/abs-2605-13083} & \rankone{0.7095} & \rankone{0.1646} & \rankone{0.1357} & \rankone{6.5846} \\
        \bottomrule
    \end{tabular*}
\end{table}
\subsubsection{Tactile from Vision}
Visual observations provide only indirect evidence of physical contact.
This ambiguity becomes particularly severe in egocentric views, where the manipulating hand frequently occludes the fingertips and the contacted object regions precisely when pressure is applied.
 
\paragraph{Task.} 
Given an egocentric video of a human-object interaction, the goal is to predict full-hand grasp pressure at each moment. 
We evaluate the predicted pressure distributions against synchronized measurements captured by the tactile glove.
 
\paragraph{Metrics.} 
We report four complementary metrics. 
Temporal accuracy measures frame-wise contact-state prediction over the interaction sequence.
Contact IoU (C-IoU) evaluates the spatial overlap between thresholded predicted and ground-truth contact regions, while volumetric IoU (V-IoU) additionally accounts for pressure magnitude through min-max aggregation.
Center-of-Pressure (CoP) error measures the spatial distance between predicted and ground-truth pressure centroids over the fingertips and palm, with lower values indicating more accurate contact localization.
 
\paragraph{Baselines.} We evaluate three baseline methods using their released pre-trained weights. PressureVision~\cite{DBLP:conf/eccv/GradyTBTWHK22} directly regresses pixel-aligned pressure maps from RGB observations using a convolutional encoder-decoder. 
EgoPressureDiff~\cite{DBLP:conf/icml/abs-2606-09243} formulates pressure estimation as conditional diffusion and adapts a pre-trained video diffusion backbone. 
TouchAnything~\cite{DBLP:journals/corr/abs-2605-13083} learns a general vision-to-touch representation using cross-view fusion and view-dropout training.
 
\paragraph{Results.}
Table~\ref{tab:tactile} reports the results on the close-range table-scale recordings. 
PressureVision provides little meaningful contact prediction, producing nearly zero overlap under both C-IoU and V-IoU and the largest CoP error. 
EgoPressureDiff substantially improves temporal contact recognition and pressure localization, but its spatial overlap remains limited, indicating that detecting when contact occurs is considerably easier than recovering where pressure is distributed across the hand.
In contrast, TouchAnything consistently achieves the strongest performance across all four metrics.
Its advantage is especially pronounced in temporal accuracy and spatial overlap, suggesting that broader visual-tactile training and cross-view modeling improve transfer to the diverse interactions in ACE-Data-0. 
Nevertheless, its absolute C-IoU and V-IoU remain modest, and the remaining CoP error indicates that accurate pressure localization is still challenging.
In particular, a model may correctly identify the presence of contact while failing to recover its precise position and intensity under hand-object occlusion.

Overall, the results reveal a substantial generalization gap for existing tactile-estimation models and establish ego-view pressure reconstruction as a challenging benchmark. 
Robust performance requires not only recognizing contact events, but also resolving fine-grained pressure distributions from visually ambiguous and frequently occluded interactions.

\subsection{Scene Components}
\label{sec:bench_scene}
Where the previous track asks what an interaction feels like, this one asks where the components of the scene are and how they move over time. Between raw sensor signals and any understanding of an interaction lie the 3D states of the scene: the pose of the body, the articulation of the hands, and the 6-DoF pose of every object. These states are what most vision pipelines estimate first and what every downstream module consumes.

In home scenes, however, their reliability remains largely unquantified. 
In-the-wild footage offers no ground truth to measure against, datasets with metric ground truth are confined to laboratory settings, and datasets recorded in real environments are pseudo-labeled by the very methods under evaluation.
ACE-Data-0 removes this obstacle by providing metric ground truth in furnished domestic scenes, allowing existing estimators to be evaluated under the conditions in which they are actually deployed. 
We therefore benchmark human motion estimation in this track. Hand articulation is treated separately in the following track, where dexterous manipulation provides the setting in which it matters most. Object pose estimation is left to future work: too few applicable methods~\cite{wen2023foundationpose} exist to support a meaningful comparison, though the released ground truth supports this evaluation directly.

\begin{table}[t]
\centering
\caption{\textbf{Human motion estimation on ACE-Data-0.} All metrics in mm.}
\label{tab:human_motion}
\footnotesize
\setlength{\tabcolsep}{6pt}
\begin{tabular*}{\linewidth}{@{\extracolsep{\fill}}l c c c c c}
\toprule
\textbf{Method} & \textbf{PA-MPJPE} $\downarrow$ & \textbf{PA-PVE} $\downarrow$ & \textbf{MPJPE} $\downarrow$ & \textbf{PVE} $\downarrow$ & \textbf{WA-MPJPE} $\downarrow$ \\
\midrule
\rowcolor{black!6}
\multicolumn{6}{@{}l}{\textit{\small\textcolor{black!65}{--- Single-view exocentric, per-frame ---}}} \\
\noalign{\vspace{3pt}}
Multi-HMR~\cite{baradel2024multihmr} & 110.5 & -- & 115.3 & -- & -- \\
Multi-HMR2~\cite{fiche2026multihmr2} & 85.5 & -- & 92.8 & -- & -- \\
SAM-3D-Body~\cite{yang2026sam3dbody} & 71.0 & -- & 76.4 & -- & -- \\
PyMAF-X~\cite{zhang2022pymafx} & 94.7 & 177.5 & 99.9 & 177.0 & -- \\
PARE~\cite{kocabas2021pare} & 98.7 & 132.3 & 102.0 & 134.8 & -- \\
CameraHMR~\cite{patel2024camerahmr} & 78.7 & 98.7 & 85.7 & 105.2 & -- \\ 
OSX~\cite{lin2023osx} & \rankthree{59.2} & \rankthree{73.7} & \rankthree{62.5} & \rankthree{76.6} & -- \\
\midrule
\rowcolor{black!6}
\multicolumn{6}{@{}l}{\textit{\small\textcolor{black!65}{--- Single-view exocentric, temporal ---}}} \\
\noalign{\vspace{3pt}}
SMPLer-X~\cite{cai2023smplerx} & \ranktwo{57.0} & \ranktwo{70.2} & \ranktwo{59.8} & \ranktwo{71.7} & -- \\
SMPLest-X~\cite{yin2025smplestx} & \rankone{55.7} & \rankone{65.8} & \rankone{58.8} & \rankone{67.6} & -- \\
% SAM-Body4D~\cite{gao2025sambody4d} & & & & & \\
Humans-in-4D~\cite{goel2023humans4d} & 77.1 & 94.4 & 79.6 & 96.4 & -- \\
GVHMR~\cite{Shen_2024worldground} & 88.0 & 104.1 & 96.3 & 112.0 & \rankthree{217.1} \\
WHAM~\cite{DBLP:conf/cvpr/Shin0HB24} & 131.1 & 160.2 & 144.0 & 166.4 & 243.1 \\
EasyMoCap~\cite{easymocap, shuai2022multinb} & 103.4 & 155.8 & 106.0 & 156.1 & 256.2\\
\midrule
\rowcolor{black!6}
\multicolumn{6}{@{}l}{\textit{\small\textcolor{black!65}{--- Single-view exocentric, scene-aware ---}}} \\
\noalign{\vspace{3pt}}
Phy-SIC~\cite{muralidhar2025physic} & 64.3 & 78.9 & 71.3 & 84.6 & -- \\
UniSH~\cite{li2026unish} & 80.0 & 111.2 & 82.8 & 114.0 & \ranktwo{196.3} \\
JOSH~\cite{liu2025josh} & 64.0 & 89.2 & 69.9 & 95.0 & 245.1 \\
Human3R~\cite{chen2026human3r} & 60.1 & 75.4 & 63.5 & 80.6 & \rankone{180.2} \\
\midrule
\rowcolor{black!6}
\multicolumn{6}{@{}l}{\textit{\small\textcolor{black!65}{--- Multi-view exocentric ---}}} \\
\noalign{\vspace{3pt}}
MAMMA~\cite{cuevasvelasquez2025mamma} & 70.1 & 86.8 & 74.8 & 89.6 & 230.8 \\
U-HMR~\cite{li2024uhmr} & 134.4 & 176.0 & 150.2 & 179.2 & -- \\
HSfM~\cite{muller2024hsfm} & 92.8 & 133.8 & 93.8 & 134.8 & -- \\
\midrule
\rowcolor{black!6}
\multicolumn{6}{@{}l}{\textit{\small\textcolor{black!65}{--- Egocentric ---}}} \\
\noalign{\vspace{3pt}}
EgoEgo~\cite{li2022egoego} & 159.6 & 245.3 & 163.4 & 263.9 & 306.2 \\
EgoAllo~\cite{yi2024egoallo} & 131.7 & 196.8 & 147.9 & 220.3 & 252.2 \\
\bottomrule
\end{tabular*}
\end{table}
\subsubsection{Human Motion Estimation}
Human motion estimation in home environments remains challenging for several reasons. 
Furniture often blocks the lower body for long parts of a sequence.
Common household actions, such as crouching in front of a cabinet, reaching above the head, or lying on a sofa, also differ from the mostly upright poses found in many training datasets. 
In addition, each take can last several minutes, allowing small frame-level errors to build up and cause large drift in the estimated global trajectory~\cite{zhang2022egobody,DBLP:conf/cvpr/Shin0HB24,Shen_2024worldground,muralidhar2025physic}.
Our dataset captures these challenges together and provides metric ground-truth motion for the full sequence.

\paragraph{Task.}
Given the visual observations from a take, the goal is to estimate the articulated body pose over time. 
Per-frame methods process individual images, while temporal methods use the full video sequence. 
Methods with world-frame outputs are also expected to recover the person's global trajectory through the scene.

We evaluate all predictions against motion-captured ground truth. 
The evaluation covers three input settings: multi-view exocentric video, single-view exocentric video, and egocentric video.
It also includes three method families: per-frame, temporal, and scene-aware methods.
Since all settings use the same ground truth, the results allow us to compare the effects of viewpoint, temporal information, and scene context directly.

\paragraph{Metric.}
We report five metrics, ordered from local to global.
PA-MPJPE and PA-PVE~\cite{ionescu2014human36m} evaluate the articulated pose and the recovered body surface after Procrustes alignment, independent of global position and orientation.
MPJPE and PVE~\cite{loper2015smpl,pavlakos2019expressivebodycapture3d} evaluate the same quantities with the global orientation retained. 
WA-MPJPE~\cite{DBLP:conf/cvpr/Shin0HB24,Shen_2024worldground} evaluates the global trajectory in the world frame, after a single alignment of the whole path.

\paragraph{Baselines.}
We evaluate 22 methods, covering every major family of human motion estimation. 
For single-view exocentric input, these include per-frame approaches (Multi-HMR~\cite{baradel2024multihmr}, Multi-HMR2~\cite{fiche2026multihmr2}, SAM-3D-Body~\cite{yang2026sam3dbody}, PyMAF-X~\cite{zhang2022pymafx}, PARE~\cite{kocabas2021pare}, CameraHMR~\cite{patel2024camerahmr}, and OSX~\cite{lin2023osx}), video-based approaches (SMPLer-X~\cite{cai2023smplerx}, SMPLest-X~\cite{yin2025smplestx}, Humans-in-4D~\cite{goel2023humans4d}, GVHMR~\cite{Shen_2024worldground}, WHAM~\cite{DBLP:conf/cvpr/Shin0HB24}, and EasyMoCap~\cite{shuai2022multinb}), and their scene-aware counterparts, Phy-SIC~\cite{muralidhar2025physic} for the former and UniSH~\cite{li2026unish}, JOSH~\cite{liu2025josh}, and Human3R~\cite{chen2026human3r} for the latter. 
For multi-view exocentric input, we evaluate MAMMA~\cite{cuevasvelasquez2025mamma}, U-HMR~\cite{li2024uhmr}, and HSfM~\cite{muller2024hsfm}. 
For egocentric input, we evaluate EgoEgo~\cite{li2022egoego} and EgoAllo~\cite{yi2024egoallo}. 
All methods run with their released pre-trained weights. 
To our knowledge, this is the broadest evaluation of human motion estimation conducted in real home scenes to date.

\paragraph{Results.}
We evaluate primarily on the room-scale recordings, which combine locomotion, furniture occlusion, and minutes of continuous motion.
Quantitative results are reported in Table~\ref{tab:human_motion}, and three findings stand out.

First, the results show a clear gap between local pose estimation and global trajectory recovery.
Several methods achieve strong results on the Procrustes-aligned metrics, with accuracy close to that reported on standard benchmarks. This suggests that they can still recover articulated body poses under household occlusion and uncommon postures.
However, their world-frame trajectory errors remain much higher. 
The ranking on local pose metrics also differs from that on global trajectory metrics. 
In other words, estimating the body pose correctly in each frame does not guarantee an accurate motion path over the full sequence.

Second, the comparison between scene-aware and temporal methods further shows where scene information is most useful.
Scene-aware methods achieve lower trajectory errors, while their Procrustes-aligned results remain similar to those of temporal methods.
This suggests that scene context mainly helps determine the body's position in the room, rather than improving the relative joint configuration. 
Scene geometry provides useful constraints on where a person can stand or move, but gives less direct information about the exact body pose.

Third, the results also show a strong effect of viewpoint.
Egocentric methods perform worse across most metrics because much of the body is outside the camera's field of view and must be inferred from head motion. 
Multi-view methods, however, do not outperform the strongest single-view methods in this evaluation.
We do not view this as evidence that multiple views are less useful.
Instead, it likely reflects the strong performance of recent single-view whole-body and scene-aware methods, as well as the limited range of available multi-view baselines.

\subsection{Embodied Interaction}
\label{sec:bench_interaction}
Knowing the position of the person is not enough to describe how an interaction is carried out. 
Much of this information is contained in the hand motion, including how the hand approaches an object, forms a grasp, adjusts its position, and releases it.
Hand motion is also especially relevant to robot learning because a hand trajectory can be retargeted to a robotic gripper more directly than raw visual observations.

We therefore evaluate how accurately current methods~\cite{qin2022dexmv,kareer2025egomimic,punamiya2025egobridge} recover hand motion from video. 
In ACE-Data-0, each interaction is recorded at the same time from both egocentric and exocentric viewpoints. 
This allows us to compare the two settings under the same conditions, using the same takes and the same ground truth while changing only the viewpoint.

\begin{table}[t]
    \centering
    \caption{\textbf{Hand motion estimation from ego-view on ACE-Data-0}.}
    \label{tab:hoi_ego}
    \footnotesize
    \setlength{\tabcolsep}{6pt}
    \begin{tabular*}{\linewidth}{@{\extracolsep{\fill}}l c c c c c c}
        \toprule
        \textbf{Method} & \textbf{PA-MPJPE} $\downarrow$ & \textbf{MPJPE} $\downarrow$ & \textbf{F@5} $\uparrow$ & \textbf{F@15} $\uparrow$ & \textbf{AUC$_J$} $\uparrow$ & \textbf{Traj.\ err.} $\downarrow$ \\
        \midrule
        WildHands~\cite{prakash20243d}  & \rankone{11.2} & \rankone{12.6} & \rankone{0.175} & \rankone{0.774} & \rankone{0.776} & --\\
        Dyn-HaMR~\cite{yu2024dynhamr} & 18.9 & 21.1 & 0.042 & 0.413 & 0.624 & \rankone{98.2}\\
        HaWoR~\cite{zhang2025hawor} & 13.8 & 17.4 & 0.130 & 0.666 & 0.729 & 102.1\\
        \bottomrule
    \end{tabular*}
\end{table}
\subsubsection{HOI from Ego-View}
\label{sec:hoi_ego}
Egocentric video provides the viewpoint that is most similar to what a future robot may observe. 
The camera stays close to the hands and often captures fine details of the interaction. 
However, this viewpoint also introduces several practical challenges.
The camera moves with the head, the wide-angle lens distorts the image near the boundary, and the hands often appear in these distorted regions.
Large head or body movements may also move one or both hands outside the field of view.

\paragraph{Task.} 
Given the egocentric video of a take, a method estimates the 3D motion of both hands as MANO~\cite{Romero_2017} sequences.
This includes the articulated finger pose in each frame and, when supported by the method, the hand trajectory in a world coordinate frame.
We obtain the ground truth from the Manus motion-capture gloves for the room-scale recordings; for the table-scale recordings, we apply RANSAC triangulation to 2D hand keypoints from the eight synchronized exocentric cameras, followed by manual refinement.
Joints that are visible in too few views are masked during evaluation. 
The method is therefore not penalized on frames for which reliable ground truth cannot be obtained.
 
\paragraph{Metrics.}
We report five metrics for hand pose and one metric for the global trajectory. 
The same definitions and units are used for both egocentric and exocentric evaluation.
PA-MPJPE applies a separate Procrustes alignment to each predicted hand using rotation, translation, and scale. 
It therefore mainly measures errors in finger articulation.
MPJPE uses only rotation and translation while keeping the scale fixed, and thus also reflects errors in metric hand size.
F@5 and F@15 report the fractions of joints with errors below 5 and 15 mm, respectively.
AUC$_J$ is the normalized area under the PCK curve over thresholds from 0 to 50 mm.
For methods that predict a persistent world coordinate frame, we also report the world-frame trajectory error. 
We align the predicted and ground-truth wrist trajectories using one similarity transform for the entire clip and compute the mean residual error.
Unlike per-frame alignment, this metric captures drift and inconsistent scale across the sequence.
 
\paragraph{Baselines.} 
We evaluate three methods using their released pre-trained weights.
WildHands~\cite{prakash20243d} is a per-frame regressor and is applied independently to each egocentric frame. 
Dyn-HaMR~\cite{yu2024dynhamr} and HaWoR~\cite{zhang2025hawor} process videos and recover the hands in a world coordinate frame by combining hand estimation with camera-motion estimation.

\paragraph{Results.} 
Table~\ref{tab:hoi_ego} shows a clear difference between frame-level hand pose estimation and world-space hand motion recovery. 
The per-frame method achieves the strongest articulation accuracy, while the two video-based methods perform less well on the local pose metrics. 
One possible reason is that the video-based methods estimate camera motion and hand pose together.
Errors in camera motion can therefore affect the recovered hand pose, especially in our recordings where head motion is often large.
The difference becomes more evident for global trajectory estimation.
Both world-space methods produce trajectory errors that are much larger than their local joint errors. 
This suggests that the main challenge in egocentric hand reconstruction is not estimating finger articulation in individual frames, but maintaining a stable hand trajectory in the world coordinate system. 
Together with the exocentric results below, this points to egomotion estimation as a major source of error.

\begin{table}[t]
    \centering
    \caption{\textbf{Hand motion estimation from exo-view on ACE-Data-0}.}
    \label{tab:hoi_exo}
    \footnotesize
    \setlength{\tabcolsep}{6pt}
    \begin{tabular*}{\linewidth}{@{\extracolsep{\fill}}l c c c c c c}
        \toprule
        \textbf{Method} & \textbf{PA-MPJPE} $\downarrow$ & \textbf{MPJPE} $\downarrow$ & \textbf{F@5} $\uparrow$ & \textbf{F@15} $\uparrow$ & \textbf{AUC$_J$} $\uparrow$ & \textbf{Traj.\ err.} $\downarrow$ \\
        \midrule
        \noalign{\vspace{3pt}}
        HORT~\cite{chen2025hort} & 10.8 & 12.3 & \rankthree{0.276} & 0.776 & 0.784 & -- \\
        HaMeR~\cite{pavlakos2023hamer} & \ranktwo{9.6} & \ranktwo{10.4} & \ranktwo{0.281} & \ranktwo{0.848} & \ranktwo{0.812} & -- \\
        HaPTIC~\cite{ye2025haptic} & \rankthree{10.0} & \rankthree{10.7} & 0.247 & \rankthree{0.838} & \rankthree{0.804} & \rankone{63.0} \\
        WiLoR~\cite{potamias2024wilor} & \rankone{9.1} & \rankone{9.9} & \rankone{0.313} & \rankone{0.854} & \rankone{0.819} & -- \\
        OmniHands~\cite{lin2024omnihands} & 10.7 & 11.5 & 0.211 & 0.814 & 0.791 & -- \\
        \bottomrule
    \end{tabular*}
\end{table}
\subsubsection{HOI from Exo-View}
Exocentric video presents a different set of advantages and challenges. 
The camera remains fixed, and the person and surrounding scene usually stay within the image. 
This provides a stable reference frame for tracking motion over time.
However, the hands occupy only a small part of the image and are often occluded by the manipulated object or by the person's body~\cite{fu2025gigahands,fan2023arctic}. 
Since the egocentric and exocentric cameras record the same interactions, we can compare the two viewpoints using the same takes and ground truth.
 
\paragraph{Task.} 
Given either one exocentric stream or all synchronized exocentric streams, a method estimates the articulated pose of both hands in each frame.
Methods with world-frame outputs additionally recover the hand trajectories across the sequence. 
The ground truth and evaluation procedure are the same as those used in Section~\ref{sec:hoi_ego}, allowing direct comparison between the two viewpoints.
 
\paragraph{Metrics.} 
We use the same five hand-pose metrics defined in Section~\ref{sec:hoi_ego}. 
For video methods, we also report the world-frame trajectory error.
Although the exocentric camera is static, a method must still recover hand positions consistently and at the correct metric scale throughout the clip.
 
\paragraph{Baselines.} 
We evaluate five single-view methods using their released pre-trained weights. 
HaMeR~\cite{pavlakos2023hamer}, WiLoR~\cite{potamias2024wilor}, and OmniHands~\cite{lin2024omnihands} estimate the hands independently in each frame. 
HaPTIC~\cite{ye2025haptic} processes video and produces hand motion in a world coordinate frame. 
HORT~\cite{chen2025hort} jointly reconstructs the right hand and the manipulated object, and therefore produces predictions only for frames containing object manipulation.
 
\paragraph{Results.} 
Table~\ref{tab:hoi_exo} summarizes the results. The five methods achieve similar articulation accuracy, with PA-MPJPE values ranging from 9.1 to 10.8 mm.
WiLoR performs best, reaching 9.1 mm PA-MPJPE, an F@5 score of 0.313, and an AUC$_J$ of 0.819. 
HaMeR follows closely with a PA-MPJPE of 9.6 mm. 
These results indicate that current crop-based hand regressors can recover detailed hand poses even when the hands occupy a small region of a 1080p image.

The fixed camera also makes global trajectory estimation more reliable. 
HaPTIC obtains a trajectory error of 63 mm while maintaining a PA-MPJPE of 10.0 mm, which is comparable to the per-frame methods.
Its trajectory error is substantially lower than the 98--102 mm obtained by the egocentric world-space methods.
This difference suggests that a stable camera coordinate system removes much of the uncertainty caused by egomotion.

HORT achieves a PA-MPJPE of 10.8 mm on the manipulation frames for which it produces predictions. 
This result is close to those of the other methods. 
Under this evaluation, jointly modeling the held object does not provide a clear improvement in articulated hand-pose accuracy, but it also does not noticeably reduce it.

\subsubsection{Cross-View Analysis}
Because the two viewpoints observe the same interactions, their results can be compared directly. 
Although the egocentric camera provides a closer view of the hands, the exocentric methods achieve better results for both articulation and global trajectory estimation.
Among these results, the trajectory analysis shows the largest difference. 
The exocentric video method obtains an error of 63 mm, while the egocentric methods remain close to 100 mm.
With a fixed exocentric camera, the camera coordinate system already provides a stable reference frame. 
Egocentric methods must estimate this reference frame from head motion, and errors in this step become a major part of the final trajectory error.

Overall, the two viewpoints still provide complementary observations. 
Egocentric video is more affected by truncation, distortion, and motion blur, while exocentric video is more affected by occlusion from the body and manipulated objects.
Combining both streams could therefore reduce their individual failure cases. 
Providing measured headset motion to egocentric methods would also help separate errors caused by hand reconstruction from those caused by egomotion estimation.
In addition, supplying object pose as an auxiliary input could test whether scene and object information improve the recovery of grasps and hand motion.
We believe that ACE-Data-0 will support further research on combining egocentric and exocentric views for more accurate pose estimation.

\section{Conclusion}

We presented ACE, an ambient capture methodology for recording everyday household behavior in real homes. ACE uses two complementary configurations at different spatial scales: a table-scale setup for fine-grained hand-object manipulation and a room-scale setup for whole-body activity across a fully furnished apartment. Both capture synchronized egocentric and exocentric video, body, hand, and object motion, audio, and tactile signals. An optical-clock procedure aligns all streams to the motion-capture clock at millisecond precision, while marker-bridged calibration registers static and wearable cameras in a common world frame.

Using ACE, we collected ACE-Data-0, comprising over 150 hours of recording, 75,000 interaction episodes, 17M frames, and per-frame annotations. We benchmarked existing methods on touch prediction from video, full-body motion recovery, and hand-motion estimation from egocentric and exocentric views. Across these tasks, existing methods degrade under contact, occlusion, and long-duration activity---conditions that distinguish real homes from controlled laboratories. These results motivate models that fuse views and modalities, enforce physical constraints, and learn from measured contact and motion supervision.

By combining egocentric observations, demonstration trajectories, and contact-level supervision in a single time-aligned stream, ACE-Data-0 is designed to support research on manipulation policies, world models, and vision-language-action systems that connect perception, action, and physical state in real homes.

\paragraph{Limitations.}
ACE has several limitations. First, it covers only two sites and therefore captures limited variation in layouts, furnishings, and lighting. Second, ground truth is restricted to instrumented entities: tracked objects must be scanned and equipped with markers in advance, and the dataset does not annotate state changes of articulated mechanisms, fluids, or deformable materials. Third, the suit, gloves, headset, and markers remain visible in the recordings and may introduce dataset-specific visual cues.

\paragraph{Ethics statement.}
All participants volunteered and provided informed consent for both recording and public data release.

\clearpage
\bibliographystyle{unsrtnat}
\bibliography{references}

\end{document}